\documentclass[10pt,twocolumn,letterpaper]{article}

\usepackage[pagenumbers,algorithms]{wacv}      
\usepackage{graphicx}%
\usepackage{multirow}%
\usepackage{amsmath,amssymb,amsfonts}%
\usepackage{amsthm}%
\usepackage{mathrsfs}%
\usepackage[title]{appendix}%
\usepackage{xcolor}%
\usepackage{textcomp}%
\usepackage{manyfoot}%
\usepackage{booktabs}%
\usepackage{algorithm}%
\usepackage{algorithmicx}%
\usepackage{algpseudocode}%
\usepackage{listings}
\usepackage{graphicx}
\usepackage{booktabs}
\usepackage[misc]{ifsym}
\usepackage{mwe}
\usepackage{ulem}

\newcommand{\val}[2]{#1{\tiny{$\pm$#2}}}
\newcommand{\valb}[2]{$\mathbf{#1} {\scriptstyle \, \pm \, #2}$}

\newcommand{\pgo}{PGO-BEn}

\definecolor{wacvblue}{rgb}{0.21,0.49,0.74}
\usepackage[pagebackref,breaklinks,colorlinks,allcolors=wacvblue]{hyperref}

\def\wacvPaperID{*****} 
\def\confName{WACV}
\def\confYear{2027}

\definecolor{wacvblue}{rgb}{0.21,0.49,0.74}
\usepackage[pagebackref,breaklinks,colorlinks,allcolors=wacvblue]{hyperref}

\def\wacvPaperID{1} 
\def\confName{A}
\def\confYear{2027}

\title{Few-Shot Domain Incremental Learning via Continual Vision-Language Consolidation}

\author{Naeem Paeedeh$^{1}$ \quad
Mahardhika Pratama$^{1}$ \quad
Wolfgang Mayer$^{1}$ \quad
Mukesh Prasad$^{2}$ \\
Weiping Ding$^{3}$ \quad
Yew-Soon Ong$^{4}$ \\
\\
$^{1}$School of Computer Science and IT, Adelaide University, Australia. \\
$^{2}$Faculty of Engineering and IT, University of Technology Sydney (UTS), Australia \\
$^{3}$School of Artificial Intelligence and Computer Science, Nantong University, China \\
$^{4}$College of Computing and Data Science, Nanyang Technological University, Singapore \\
{\tt\small \{naeem.paeedeh, dhika.pratama, wolfgang.mayer\}@adelaide.edu.au}, \\
\tt\small mukesh.prasad@uts.edu.au, dwp9988@163.com, asysong@ntu.edu.sg
}

\begin{document}
\maketitle
\begin{abstract}
Existing domain-incremental learning (DIL) strategies call for massive amounts of data to adapt to new domains and suffer from the overfitting problem in the case of data scarcity. This paper puts forward a relatively uncharted problem, namely, few-shot domain incremental learning (FSDIL), taking into account the problem of extreme data shortages in the realm of DIL. A novel algorithm, namely Continual Vision-Language Consolidation (CVLC), is proposed to address the FSDIL problem, where the key idea lies in the concept of latent space reservation in the base domain coupled with dual coalescent projection (DCP) as a parameter-efficient fine-tuning method. First, the vision prototype is calibrated while multiple templates and synonyms are generated via LLMs to induce the language prototype. The vision and language prototypes are fused. Adaptation to never-ending arrivals of new domains is done by the DCP technique, fine-tuned in such a way to prepare the model to unseen domains via latent-space reservations committed in the base domain. CVLC is structured under shared and domain-specific components to combine general knowledge and domain-specific details. The advantage of our approach is demonstrated through a range of benchmark problems and comparisons with prior arts, in which CVLC outperforms them by up to a $16\%$ gap. Our codes are shared publicly in \href{https://github.com/Naeem-Paeedeh/CVLC}{https://github.com/Naeem-Paeedeh/CVLC}. 
\end{abstract}
    
\section{Introduction}
Continual learning (CL) \cite{chen2018lifelong,Parisi2018ContinualLL,Masana2020ClassIncrementalLS,Zhou2024ContinualLW} is a growing research territory where the goal is to address dynamic and evolving learning environments. It goes beyond the conventional training paradigm where a model isn't kept fixed once deployed, but rather continuously adapted to keep pace with rapidly changing environments. The underlying challenge lies in the stability-plasticity dilemma, where a plastic model is capable of adapting to new conditions, but its old knowledge is compromised due to the catastrophic forgetting (CF) problem. On the other hand, a stable model retains past knowledge but fails to adapt to new environments. 

The CL problem can be divided into three sub-branches \cite{vandeVen2019ThreeSF}: Class-Incremental Learning (CIL) \cite{Zhou2023ClassIncrementalLA}, where a model is trained to recognize new categories, Task-Incremental Learning (TIL) relying on task identifiers (IDs) for inferences, and Domain-Incremental Learning (DIL), where a model is trained to address a sequence of varying domains. Although the CL area has progressed rapidly, most efforts are devoted to the CIL and TIL problems \cite{Zhou2023ClassIncrementalLA}. Besides, na\"ive translations of the CIL approaches to the DIL context usually lead to suboptimal performances.  

The DIL problem comprises a sequence of different domains while having a fixed set of target classes. In other words, changes are induced by domain shift problems caused by style shifts, environmental changes, data quality degradations, etc. \cite{Liu2024CompositionalPF} addresses the DIL problem with the idea of compositional prompt using domain-specific prompt pools. \cite{Wang2025DualCPRD} introduces the concept of dual-level concept prototypes consisting of coarse-grained prototypes and fine-grained prototypes. A dual consolidation approach at the feature level and the classifier level is proposed in \cite{Zhou2024DualCF}. \cite{Wang2025BoostingDI} relies on a domain ID predictor, putting forward the concept of a Gaussian mixture compressor and domain feature resampler. \cite{Paeedeh2025ContinualKC} puts the task-shared LoRAs and the task-shared LoRAs under one roof, where an auxiliary network is incorporated to identify the best task-specific component for inferences. These approaches call for a lot of samples, which can be hard to obtain in practice. 

To the best of our knowledge, \cite{mukherjee2026textscpgoben} constitutes the first and only attempt to cope with the data scarcity issue in the DIL, where it formalizes the few-shot domain-incremental learning problem (FSDIL). Nevertheless, this approach relies on the prompt tuning approach, where the prompts are injected across all layers of the vision encoder and the text encoder. This approach relies on the CLIP's conventional class name template. Such an approach is often ambiguous and insufficient to detail an object, thus resulting in inferior performance. Besides, \cite{mukherjee2026textscpgoben} ignores shareable information across domains. Related works are further detailed in \textbf{the supplemental document}.

This paper proposes a novel approach for the FSDIL problem, namely Continual Vision-Language Consolidation (CVLC), featuring the latent space reservation concept via a dual coalescent projection (DCP) tuning strategy. CVLC is built upon the CLIP backbone with the vision and text encoders where domain-shared DCP and domain-specific DCP are integrated to explore common and domain-specific knowledge. The domain-specific DCP is trained across all domains and occupies the first $l$ transformer blocks while the domain-specific DCP is assigned to accommodate domain-specific details across the last $L-l$ transformer blocks. The DCP concept constitutes a parameter-efficient fine-tuning (PEFT) approach based on only a dual learnable matrix combining the query, key and value of the self-attention layer. As a result, it is more parameter-efficient than the LoRA method \cite{zhang2026exemplar} and doesn't need to select the rank of learnable matrices.  The latent space reservation concept prepares the network in the base domain to face unseen domains via generations of imaginary classes. That is, the imaginary classes guide the model to reserve the latent space for future domains. On the other hand, a non-parametric approach is developed to predict the domain identifiers (IDs) for inferences.

Another unique feature of our approach lies in the use of linear combinations between multiple templates and its weighted synonyms, thus enriching semantics of an object leading to improved discriminative features. The vision and text prototypes are crafted and calibrated afterward to induce final predictions. Last but not least, a prototype correction strategy is incorporated to counteract the representational drift problem causing old prototypes to be obsolete. Such strategy measures the displacement of the prototypes and corrects their locations accordingly. The advantage of CVLC is demonstrated through a range of benchmarks problems and rigorous comparisons with recently published works. It is shown that CVLC outperforms prior arts with significant margins. Our major contributions are listed: 
\begin{itemize}
    \item This paper proposes a new approach for the FSDIL problem where the key idea lies in the dual coalescent projections (DCPs) trained with the latent space reservation concept in the base domain. CVLC is structured under a CLIP backbone where both visual and text encoders incorporate the domain-specific DCPs and the domain-shared DCPs. A non-parametric strategy is designed to select the domain-specific components for inferences while a prototype rectification approach is applied to address the representational drift issue.  
    \item This paper proposes a convex interpolation concept for text descriptions. Text encoder is fed with interpolation between multiple templates of the class name and its weighted synonyms to enrich semantics of an object. This leads to text prototypes further calibrated and combined with the visual prototypes to deliver the final outputs. 
    \item This paper proposes a dual coalescent projection (DCP) concept as an alternative PEFT method. This concept amalgamates the query, key and value of the self-attention layer into a single concept. It steers the direction of the attention map of the frozen transformer backbone by introducing few learnable parameters.  
    \item Rigorous experiments have been done to validate the advantage of our approach. It is shown that CVLC beats prior arts with significant margins. 
\end{itemize}
\section{Preliminaries}
\subsection{Problem Definition}
Few-Shot Domain-Incremental Learning (FSDIL) constitutes an extension of DIL that addresses data scarcity. That is, a model is presented with a sequence of unique domains $\{\mathcal{D}_{1},\mathcal{D}_{2},...,\mathcal{D}_{T}\}$ where $T$ is the number of domains unknown before the process runs. Each domain shares the same label space, where $\forall t, t^{'}\in[1,T] \mathcal{Y}_{t}=\mathcal{Y}_{t^{'}}$ yet features the domain shift problem $P(\mathcal{X},\mathcal{Y})_{t}\neq P(\mathcal{X},\mathcal{Y})_{t^{'}}$ due to style shifts, environmental changes, data quality degradations, etc. No domain identifiers (IDs) are offered for inferences. In other words, a model is supposed to predict the domain IDs independently. Because of the data scarcity issue, $\mathcal{D}_{1}=\{(x_{i},y_{i})\}_{i=1}^{N_{1}}$ is called the base domain and contains abundant samples, where $N_{1}$ denotes the cardinality of the base domain while
the remainder of domains $\mathcal{D}_{t>1}=\{(x_{i},y_{i})\}_{i=1}^{N\times K}$ is known to be an incremental domain and suffers from scarce samples formulated in the $N$ way $K$ shot setting, i.e., $N$ is the number of classes per domain, and $K$ is the number of samples per classes $N_{t}=N\times K$. $x_i\in\mathcal{X}$ is an input image and $y_{i}\in\mathcal{Y}$ is the corresponding class label. For any given domain, only data points of the most recent domains are offered. The absence of old samples results in the catastrophic forgetting (CF) problem. In addition, we focus on the exemplar-free setting, in which the use of an episodic memory $\mathcal{M}$ for storing old samples is prohibited. On the other hand, the evaluation is carried out for all already seen domains $\mathcal{D}_{1},\mathcal{D}_{2},\dots,\mathcal{D}_{t}$. 

\subsection{Vision Language Model (VLM)}
Our method, CVLC, is developed from a vision-language model (VLM), CLIP, \cite{Radford2021LearningTV}, which incorporates dual modalities, vision and text. It is built upon a vision encoder $g_{\theta_{v}}(x)\in\Re^{d}$ and a text encoder $g_{\theta_{tx}}(\omega_{k})\in\Re^{d}$ where both map each modality into a shared embedding space. $\omega_{k}$ stands for a handcrafted prompt of the $k-th$ class, e.g., "A Photo of [CLASS]". The final output of CLIP is inferred from the matching degree between the vision and text embedding:
\begin{equation}
\begin{split}
    p_{\text{VLM}}(y|x)=\frac{\exp{(\text{s}(g_{\theta_{v}}(x),g_{\theta_{tx}}(\omega_{k}))/\tau)}}{\sum_{k=1}^{M}\exp{(\text{s}(g_{\theta_{v}}(x),g_{\theta_{tx}}(\omega_{k}))/\tau)}}
\end{split}
\end{equation}
where $\text{s}(.)$ is a cosine similarity function and $\tau$ is a temperature constant. $M$ denotes the number of target classes. CLIP exhibits a zero-shot learning property and can be applied directly to the downstream tasks because it was pretrained with millions of image-text pairs.

\subsection{Prototype-based Classifier}
The prototype-based classification scheme is driven by prototypes and deemed more efficient than linear classifier because prototypes are directly determined from data samples and require no gradient computation. That is, it is less prone to the overfitting issue than the linear classifier in the case of low samples. Let $z\in\Re^{d}$ be a feature embedding, the prototypes are defined as empirical means of the target classes. 
\begin{equation}
    \mu_{m}=\frac{1}{N_{m}}\sum_{i=1}^{N_{m}}\mathbb{I}_{y_{i}=m}z_{i},
\end{equation}
where $N_{m}$ is the cardinality of the $m-th$ target class and $\mathbb{I}_{y_{i}=m}$ is an indicator function being 1 if the true class label falls into the $m-th$ target class. The predicted output of the prototype-based classifier is expressed as follows:
\begin{equation}
    p(y|x)=\frac{\exp{(-\tau \operatorname{s}(\mu_{m},z))}}{\sum_{m=1}^{M}\exp{(-\tau \operatorname{s}(\mu_{m},z))}},
\end{equation}
where $\tau$ denotes a temperature controlling the sharpness.
\section{Method}
Our approach, CVLC, is based on the VLM. Specifically, it relies on the vision encoder and the text encoder working collaboratively to solve downstream tasks. We rely on the DCP concept as a PEFT strategy, in which the first $l$ transformer blocks are assigned to domain-shared DCPs while the remainder $L-l$ transformer blocks are reserved for domain-specific DCPs. This design allows both common and domain-specific knowledge to be extracted. Unlike prompt \cite{Wang2021LearningTP}, DCP combines the key, query, and value into a single concept via a dual learnable matrix. As a result, it doesn't expand the feature dimension. Also, there is no need to guess the optimum prompt length. We further extend the CLIP architecture, in which the text encoder is linked to a single and static template. Such a template is far from sufficient to capture an object's semantics. Hence, we propose the concept of convex combinations between the class name and its corresponding synonyms. That is, it applies multiple templates and synonyms of an object, thus enriching its semantics and enhancing its discriminative power. Besides, the latent-space reservation concept is implemented in the base domain to regularize the network for unseen domains. This aspect relies on generations of imaginary classes guiding the network for unseen domains. The prototype correction strategy is implemented to address the representational drift. That is, the representational drift renders previous prototypes obsolete, thus leading to the CF problem. A compensation is applied to previous prototypes when learning a new domain.    

\subsection{Dual Coalescent Projections}
A parameter-efficient fine-tuning (PEFT) approach, namely dual coalescent projection (DCP), is proposed to fine-tune a foundation model for downstream tasks without overfitting on very few samples. The key idea is to control the attention calculation connecting the query, key, and value projections using a dual learnable matrix, termed the coalescent matrix. 
\begin{equation}
    A=\operatorname{Softmax}(\frac{QC_{1}K^{T}}{\sqrt{d_{k}}})VC_{2},
\end{equation}
where $C_{1}$ is the first coalescent matrix connecting the query and key into a single concept. Since the query and key projections are frozen during the fine-tuning process to prevent the overfitting problem, the only feasible approach to steer the attention map is to insert the coalescent matrix. $C_{2}$ is the second coalescent matrix, where the main task is to steer the projections between the attention map and the value matrix.
 We simply initialize the coalescent matrix as an identity matrix in practice $C_{1},C_{2}\approx I$. Compared to the soft prompt approach \cite{Wang2021LearningTP}, a popular PEFT method, the DCP method has no user-defined parameters, whereas the prompt technique requires selecting the problem-specific prompt length. The DCP approach has the ability to handle each head separately, whereas the prompt technique shares its tokens across all heads, causing interference. Furthermore, with close to an identity matrix initialization, it does not change the behavior of the network at the beginning. Last but not least, the prompt method expands the dimensions of attention maps, whereas the DCP keeps them stable.  Compared to LoRA \cite{zhang2026exemplar,Paeedeh2025ContinualKC}, DCP is more parameter-efficient than LoRA and doesn't need to guess the rank of up- and down-projection matrices.

CVLC employs the domain-shared DCPs $\{C_{s,1}^{i},C_{s,2}^{i}\}$ for the first $l$ layers and the domain-specific DCPs $\{C_{sp,1}^{i,t},C_{sp,2}^{i,t}\}$ for the remainder of $L-l$ transformer blocks to capture both general and specific knowledge. The attention calculation is written:
 \begin{equation}
     A_{i}=\begin{cases}
         \operatorname{Softmax}(\frac{Q_{i}C_{s,1}^{i}K_{i}^{T}}{\sqrt{d_{k}^{i}}})V_{l}C_{s,2}^{i}, i\leq l\\ \operatorname{Softmax}(\frac{Q_{i}C_{sp,1}^{i,t}K_{i}^{T}}{\sqrt{d_{k}^{i}}})V_{l}C_{sp,2}^{i,t}, l < i \leq L
     \end{cases},
 \end{equation}
where $A_{i}$ stands for the output of the $i-th$ attention layer.
In other words, the shared DCPs $C_{s,1}^{i},C_{s,2}^{i}$ in early layers $i\leq l$ are devoted to extract common knowledge for all tasks while the domain-specific DCPs, $C_{sp,1}^{i,t},C_{sp,2}^{i,t}$ in deep layers $l < i \leq L$ are fine-tuned within their corresponding domain $\mathcal{D}_{t}$ to absorb the domain-specific details. 

\subsection{Textual Prototype Generations and Calibrations}
We enhance the original text encoder using the concept of multi-templates and synonyms. It is found that the original CLIP is sensitive to the template structure, and the use of multiple templates generally results in improved performance from our empirical investigation. The following 4 templates are applied in CVLC. 

\begin{align}
    \omega_{k}^{1}&=\text{"\{class\_name\}"} \\
    \omega_{k}^{2}&=\text{"\{class\_name\} .."} \\
    \omega_{k}^{3}&=\text{"This is a good \{class\_name\} .."} \\
    \omega_{k}^{4}&=\text{"It is about the \{class\_name\} .."}
\end{align}

A simple averaging method is applied across the four templates $\tilde{g}_{\theta_{tx}}(\omega_{k})=\sum_{i=1}^{P}\frac{g_{\theta_{tx}}(\omega_{k}^{p})}{P}$, where $P$ is the number of templates. The use of multiple templates doesn't suffice because it still depends on a single semantic, i.e., [Class\_name]. The concept of synonyms is introduced here to detail the semantics of an object. Specifically, we perform convex combinations between the class name and its corresponding synonyms as follows:
\begin{equation}\label{text_prototypes}
    \mu_{tx}^{k}=(1-\lambda_{tx})\tilde{g}_{\theta_{tx}}(\omega_{k})+\lambda_{tx}\sum_{o=1}^{O}s_{o,k}\tilde{g}_{\theta_{tx}}(\omega_{o,k}),
\end{equation}
\begin{equation}
   s_{o,k}=\frac{\exp{(\tau \operatorname{sim}(\tilde{g}_{\theta_{tx}}(\omega_{k}),\tilde{g}_{\theta_{tx}}(\omega_{o,k}))})}{\sum_{o=1}^{O}\exp{(\tau \operatorname{sim}(\tilde{g}_{\theta_{tx}}(\omega_{k}),\tilde{g}_{\theta_{tx}}(\omega_{o,k}))})},
\end{equation}
where $O$ denotes the number of synonyms and $\lambda_{tx}$ stands for the interpolation coefficient controlling the influence of the class name and its corresponding synonyms. $s_{o,k}$ measures the similarity between the class name and its synonyms. It quantifies the relevance of the synonyms to the class name. The text prototype $\mu_{tx}$ is crafted from a convex combination of the class name resulting from multiple prototypes and its synonyms. In other words, additional semantics are implemented to induce the text prototypes. 

\subsection{Visual Prototype Generations and Calibrations} 
The inference step of CVLC extends the original CLIP, which has both a text encoder and a vision encoder. Since our approach is inspired by CLIP \cite{Radford2021LearningTV}, it relies on text and the vision prototypes. The text prototypes are determined from \eqref{text_prototypes} using the concept of multi-templates and synonyms, while the vision prototypes are estimated from empirical means of each class as per the prototypical network \cite{Snell2017PrototypicalNF} as follows:
\begin{equation}
    \mu_{v}^{k}=\frac{1}{N_{k}}\sum_{i=1}^{N_{k}}\mathbb{I}_{y_{i}=k}z_{i},
\end{equation}
where $N_{k}$ is the cardinality of the $k-th$ class and $z_{i}$ is the vision embedding of the $i-th$ sample. Nevertheless, this approach is biased in the case of low samples as the case of the FSDIL problem, i.e., only a few samples exist when $t>1$. Calibration \cite{Chen2025EnhancingFC} is performed utilizing the base domain samples where sufficient samples are available. That is, a linear interpolation is carried out for $t>1$. 
\begin{equation}
    \tilde{\mu}_{v}^{k}=\begin{cases} \mu_{v}^{k},\quad t=1 \\ (1-\lambda_{v})\mu_{v}^{k}+\lambda_{v}\mu_{v}^{0},\quad t>1\end{cases},
\end{equation}
where $\lambda_{v}$ controls the interpolation strength between the current prototype and the base prototype. Note that the FSDIL problem features the same target classes but varying domains. This characteristic paves the way to directly interpolate the prototypes across domains.

The text prototypes and the vision prototypes operate in isolation and need to be fused to deliver improved predictions. We apply another linear interpolation to fuse the vision prototypes and the text prototypes. 
\begin{equation}
    \mu_{c}^{k}=\lambda_{c}\mu_{tx}^{k}+(1-\lambda_{c})\tilde{\mu}_{v}^{k},
\end{equation}
where $\lambda_{c}$ controls the influence of each modality. The final prediction can be calculated using the fused prototype.
\begin{equation}\label{initial}
    s_{k}=\frac{g_{\theta_{v}}(x)\mu_{c}^{k}}{|g_{\theta_{v}}(x)||\mu_{v}^{k}|},\hat{y}=\arg\max_{k=1,...,M}(s_{k})
\end{equation}
In other words, we combine the strengths of the linguistic and visual modalities. The interpolation coefficients, namely $\lambda_{tx},\lambda_{v},\lambda_{c}$, are all learned in an end-to-end fashion here rather than making them user-defined parameters. 

\subsection{Latent Space Reservations}
CVLC utilizes the latent space reservation concept \cite{chen2025pseudo}, which aims to prepare the network in the base domain for unseen future domains. The key idea is to force the model to be able to generalize beyond the original target classes. This strategy allows reservations of the latent space for the next domains. Specifically, imaginary classes are generated and learned in the base domain. That is, pseudo-classes should feature unique characteristics both to existing target classes and each other, i.e., sufficiently diverse and distinct. 

 We start with calculations of the mean and covariance matrix of each class. In other words, we assume that each class follows the normal distribution. 
\begin{equation}
    \mu_{m}=\frac{1}{N_{m}}\sum_{y_{i}=m}z_{i},
\end{equation}
\begin{equation}
    V_{m}=\frac{1}{N_{m}}\sum_{y_{i}=m}(z_{i}-\mu_{m})(z_{i}-\mu_{m})^{T},
\end{equation}
 where $\{\mu_{m},V_{m}\}$ stand for the mean and covariance matrix of the $m-th$ target classes. Inspired by \cite{verma2019manifold,chen2025pseudo}, pseudo classes are grown by linear interpolations of two classes. 
 \begin{equation}
     \mu_{\tilde{m}}=\alpha\mu_{m^{'}}+(1-\alpha)\mu_{m}
 \end{equation}
 \begin{equation}
     V_{\tilde{m}}=\alpha V_{m^{'}}+(1-\alpha)V_{m},
 \end{equation}
where $\tilde{m}$ denotes the pseudo-class candidate. $\alpha\backsim U(0,1)$ where $U$ is the uniform distribution. Here, a pool of $N_{C}$ pseudo classes are evolved $\{(\mu_{m}^{i},V_{m}^{i})\}_{i=1}^{N_{C}}$. 

Once the pseudo-classes are crafted through the interpolation method, the next question is to ensure that sufficiently diverse information is gained. That is, a filtering step is performed under two principles: 1) the pseudo-classes are sufficiently distinguishable from each other; 2) the pseudo-classes are distinct from the original target classes. Two criteria, novel-to-novel and novel-to-original, are derived.

\noindent\textbf{Novel-to-Novel Criterion}: a similarity score is calculated, which is used to prune the $N_{C}$ pseudo-classes. Let $P\in\Re^{N_{C}\times d}$ be a matrix of prototypes of pseudo-classes; the similarity matrix is defined as follows:
\begin{equation}
    S = PP^{T}
\end{equation}
\begin{equation}\label{similarity}
    S= S - (S\odot I)
\end{equation}
where $\odot$ stands for the Hadamard product. \eqref{similarity} is meant to ignore the self-similarity by zeroing the diagonal elements of the similarity matrix $S$. The similarity score is defined as the sum of each row of the similarity matrix. 
\begin{equation}
    \operatorname{Score} = S\vec{1},
\end{equation}
where $\vec{1}\in\Re^{N_{C}\times 1}$. The similarity score unveils the similarity degree between each pseudo class and another pseudo-class. $N_{0}\times N$ pseudo-classes are selected. 

\noindent\textbf{Novel-to-Original Criterion}: the second facet of the filtering process is to ensure that pseudo-classes are distinct to those of original classes. This issue can be understood by calculating the total divergence between two distributions. Since the normal distribution is assumed, the KL divergence can be implemented as follows:
\begin{equation}
\begin{split}
    D_{KL}(\tilde{m}|m)=\frac{1}{2}\sum_{m=1}^{M}[\operatorname{Tr}(V_{\tilde{m}}^{-1}V_{m})+\\(\mu_{\tilde{m}}-\mu_{m})^{T}V_{\tilde{k}}^{-1}(\mu_{\tilde{m}}-\mu_{m})+\ln{\frac{|V_{\tilde{k}}|}{|V_{k}|}}-d],
\end{split}
\end{equation}
where $\operatorname{Tr}(.)$ denotes the trace operation and $|.|$ stands for the determinant operator. $M$ is the number of original classes. We select $N$ pseudo-classes with the highest scores from the pool of $N_{C}$ candidates. Pseudo-samples can be sampled from $(\tilde{x},\tilde{y})\backsim\mathcal{N}(\mu_{m},V_{m})$ where $\tilde{y}$ denotes the interpolated label. That is, $\tilde{y}=\alpha y_{m^{'}}+(1-\alpha)y_{m}$. Such samples not only augment the sample size but also help learn diverse concepts, preventing overfitting. The latent space reservation strategy is restricted to the base domain, where sufficient samples exist. Our empirical investigation reveals that the application of this strategy in the incremental domain hinders improved generalization because it confuses the network and overlaps with those areas reserved in the base domain. In addition, incremental domains characterize very few samples and are insufficient to produce reliable statistics for linear interpolations. 

\begin{table*}[htbp]
    \centering
    \resizebox{\textwidth}{!}{
    \begin{tabular}{l c c c c c c c c c}
    \toprule
    Method & Backbone & \multicolumn{2}{c}{1-shot} & \multicolumn{2}{c}{2-shot} & \multicolumn{2}{c}{4-shot} & \multicolumn{2}{c}{8-shot} \\
    \cmidrule(lr){3-4} \cmidrule(lr){5-6} \cmidrule(lr){7-8} \cmidrule(lr){9-10}
     & & $AA^*$(↑) & $FA^*$(↑) & $AA^*$(↑) & $FA^*$(↑) & $AA^*$(↑) & $FA^*$(↑) & $AA^*$(↑) & $FA^*$(↑) \\
    \midrule
    DyTox~\cite{douillard2022dytox} & ViT & \val{58.79}{0.48} & \val{55.19}{0.96} & \val{57.91}{1.21} & \val{53.34}{0.67} & \val{55.82}{0.26} & \val{51.59}{0.84} & \val{56.18}{0.44} & \val{52.56}{0.80} \\
    LwF*~\cite{Li2016LearningWF} & CLIP & \val{62.71}{0.95} & \val{53.88}{0.56} & \val{67.82}{0.88} & \val{59.24}{0.98} & \val{71.31}{0.46} & \val{61.26}{0.87} & \val{71.16}{0.67} & \val{61.58}{0.58} \\
    EwC*~\cite{Kirkpatrick2016OvercomingCF} & CLIP & \val{63.69}{0.69} & \val{53.13}{0.38} & \val{65.32}{1.22} & \val{55.99}{0.98} & \uuline{\val{78.10}{0.87}} & \uuline{\val{70.63}{0.79}} & \val{78.09}{0.48} & \uuline{\val{69.09}{0.92}} \\
    L2P*~\cite{Wang2021LearningTP} & CLIP & \val{64.54}{0.68} & \val{58.60}{0.97} & \uuline{\val{74.84}{1.29}} & \uuline{\val{68.14}{0.69}} & \val{73.24}{0.87} & \val{65.52}{0.89} & \val{73.27}{0.93} & \val{65.43}{0.79} \\
    DualPrompt*~\cite{Wang2022DualPromptCP} & CLIP & \uuline{\val{72.12}{0.59}} & \uuline{\val{65.32}{0.67}} & \val{72.47}{0.44} & \val{65.98}{0.68} & \val{73.75}{0.88} & \val{67.44}{0.96} & \uuline{\val{75.15}{0.73}} & \val{67.33}{0.89} \\
    S-Prompt~\cite{Wang2022SPromptsLW} & CLIP & \val{63.68}{0.44} & \val{58.23}{0.37} & \val{64.23}{0.68} & \val{59.86}{0.57} & \val{65.59}{0.83} & \val{61.20}{0.86} & \val{67.74}{0.28} & \val{61.59}{0.29} \\
    CODA-Prompt~\cite{Smith2022CODAPromptCD} & CLIP & \val{71.24}{0.46} & \val{60.80}{0.67} & \val{71.23}{0.36} & \val{60.87}{0.37} & \val{70.33}{0.57} & \val{60.79}{0.28} & \val{69.34}{0.60} & \val{59.35}{0.57} \\
    InfLORA*~\cite{liang2024inflora} & CLIP & \val{62.69}{0.29} & \val{52.21}{0.83} & \val{61.58}{0.65} & \val{55.05}{0.67} & \val{68.66}{0.47} & \val{58.00}{0.66} & \val{73.70}{0.49} & \val{61.34}{0.93} \\
    CP-Prompt~\cite{feng2024cp} & CLIP & \val{66.87}{0.61} & \val{61.93}{0.36} & \val{66.94}{0.45} & \val{62.95}{0.25} & \val{66.73}{0.11} & \val{61.48}{0.17} & \val{67.26}{0.27} & \val{62.04}{0.11} \\
    \pgo~\cite{mukherjee2026textscpgoben} & CLIP & \uline{\val{74.68}{0.27}} & \uline{\val{66.34}{0.14}} & \uline{\val{77.78}{0.29}} & \uline{\val{71.84}{0.23}} & \uline{\val{83.69}{0.39}} & \uline{\val{76.79}{0.10}} & \uline{\val{84.22}{0.21}} & \uline{\val{77.71}{0.19}} \\
    \midrule
    CVLC & CLIP & \valb{87.32}{0.82} & \valb{82.04}{1.97} & \valb{87.77}{1.24} & \valb{82.36}{2.71} & \valb{88.64}{0.51} & \valb{82.78}{1.12} & \valb{89.29}{1.21} & \valb{84.55}{1.83} \\
    & $\Delta$ & +12.64 & +15.70 & +9.99 & +10.52 & +4.95 & +5.99 & +5.07 & +6.84 \\
    \bottomrule
    \end{tabular}
    }
    \caption{Comparison of different methods on the CDDB-Hard dataset. The highest values are shown in boldface, the second-highest are underlined, and the third-highest are double underlined.}
    \label{tab:cddb}
\end{table*}

\begin{table*}[htbp]
  \centering
  \label{tab:core50}
  \resizebox{\textwidth}{!}{
  \begin{tabular}{l c c c c c c c c c}
    \toprule
    Method & Backbone & \multicolumn{2}{c}{1-shot} & \multicolumn{2}{c}{2-shot} & \multicolumn{2}{c}{4-shot} & \multicolumn{2}{c}{8-shot} \\
    \cmidrule(lr){3-4} \cmidrule(lr){5-6} \cmidrule(lr){7-8} \cmidrule(lr){9-10}
     & & $AA^*$(↑) & $FA^*$(↑) & $AA^*$(↑) & $FA^*$(↑) & $AA^*$(↑) & $FA^*$(↑) & $AA^*$(↑) & $FA^*$(↑) \\
    \midrule
    DyTox~\cite{douillard2022dytox} & ViT & \val{45.06}{0.46} & \val{26.94}{1.24} & \val{47.75}{1.33} & \val{29.83}{1.13} & \val{45.14}{0.87} & \val{27.30}{0.95} & \val{48.33}{0.64} & \val{30.83}{0.48} \\
    LwF*~\cite{Li2016LearningWF} & CLIP & \val{59.40}{0.67} & \val{52.99}{0.29} & \val{66.03}{0.69} & \val{57.48}{0.58} & \val{64.61}{0.84} & \val{57.29}{0.89} & \val{67.62}{0.91} & \val{63.24}{0.63} \\
    EwC*~\cite{Kirkpatrick2016OvercomingCF} & CLIP & \val{58.43}{0.83} & \val{51.89}{0.59} & \val{66.47}{0.30} & \val{58.11}{0.41} & \val{63.55}{0.73} & \val{54.80}{0.27} & \val{65.22}{0.86} & \val{57.62}{0.47} \\
    L2P*~\cite{Wang2021LearningTP} & CLIP & \uuline{\val{80.19}{0.45}} & \uuline{\val{77.85}{0.65}} & \val{80.94}{0.96} & \uuline{\val{79.55}{0.73}} & \val{79.42}{0.63} & \val{77.98}{1.33} & \val{79.00}{0.98} & \val{78.06}{0.78} \\
    DualPrompt*~\cite{Wang2022DualPromptCP} & CLIP & \val{43.83}{0.48} & \val{36.35}{0.84} & \val{59.53}{0.45} & \val{55.28}{0.98} & \val{58.37}{0.25} & \val{52.39}{0.24} & \val{60.73}{0.84} & \val{58.16}{0.77} \\
    S-Prompt~\cite{Wang2022SPromptsLW} & CLIP & \val{77.63}{0.76} & \val{74.26}{0.57} & \val{79.53}{0.84} & \val{74.95}{0.28} & \val{79.63}{0.69} & \val{77.26}{0.58} & \val{80.13}{0.78} & \val{78.77}{0.67} \\
    CODA-Prompt~\cite{Smith2022CODAPromptCD} & CLIP & \val{53.79}{0.49} & \val{40.77}{0.87} & \val{55.09}{0.69} & \val{40.82}{0.39} & \val{57.87}{0.73} & \val{44.95}{0.87} & \val{59.97}{0.92} & \val{47.89}{0.94} \\
    InfLORA*~\cite{liang2024inflora} & CLIP & \val{57.44}{0.97} & \val{50.36}{0.77} & \val{66.47}{1.07} & \val{57.69}{0.88} & \val{63.90}{0.59} & \val{58.27}{0.47} & \val{73.12}{0.97} & \val{66.10}{0.77} \\
    CP-Prompt~\cite{feng2024cp} & CLIP & \val{78.28}{0.49} & \val{77.23}{0.36} & \uuline{\val{81.65}{0.23}} & \val{78.68}{0.56} & \uuline{\val{82.27}{1.17}} & \uuline{\val{81.26}{1.02}} & \uuline{\val{84.08}{0.67}} & \uuline{\val{82.79}{0.23}} \\
    \pgo~\cite{mukherjee2026textscpgoben} & CLIP & \uline{\val{83.19}{0.20}} & \uline{\val{78.14}{0.34}} & \uline{\val{86.73}{0.33}} & \uline{\val{82.81}{0.29}} & \uline{\val{87.38}{0.51}} & \uline{\val{83.81}{0.86}} & \uline{\val{88.43}{0.37}} & \uline{\val{85.34}{0.66}} \\
    \midrule
    CVLC & CLIP & \valb{95.74}{0.13} & \valb{94.35}{0.07} & \valb{96.21}{0.17} & \valb{95.11}{0.20} & \valb{96.44}{0.14} & \valb{95.43}{0.13} & \valb{96.96}{0.06} & \valb{96.08}{0.13} \\
    & $\Delta$ & +12.55 & +16.21 & +9.48 & +12.30 & +9.06 & +11.62 & +8.53 & +10.74 \\
    \bottomrule
  \end{tabular}
  }
  \caption{Comparison of different methods on the CORe50 dataset. The highest values are shown in boldface, the second-highest are underlined, and the third-highest are double underlined.}
\end{table*}

\begin{table*}[htbp]
\centering
\label{tab:domain_net}
\resizebox{\textwidth}{!}{%
\begin{tabular}{l c c c c c c c c c}
    \toprule
    Method & Backbone & \multicolumn{2}{c}{1-shot} & \multicolumn{2}{c}{2-shot} & \multicolumn{2}{c}{4-shot} & \multicolumn{2}{c}{8-shot} \\
    \cmidrule(lr){3-4} \cmidrule(lr){5-6} \cmidrule(lr){7-8} \cmidrule(lr){9-10}
     & & $AA^*$(↑) & $FA^*$(↑) & $AA^*$(↑) & $FA^*$(↑) & $AA^*$(↑) & $FA^*$(↑) & $AA^*$(↑) & $FA^*$(↑) \\
    \midrule
    DyTox~\cite{douillard2022dytox} & ViT & \val{29.94}{0.68} & \val{18.72}{0.59} & \val{29.20}{0.86} & \val{18.20}{0.66} & \val{35.59}{1.1} & \val{22.57}{0.62} & \val{29.71}{0.96} & \val{18.58}{0.68} \\
    LwF*~\cite{Li2016LearningWF} & CLIP & \val{72.13}{0.55} & \val{61.51}{1.20} & \val{72.26}{0.75} & \val{61.26}{0.67} & \val{72.08}{0.68} & \val{60.47}{1.21} & \val{71.77}{1.31} & \val{59.58}{0.65} \\
    EwC*~\cite{Kirkpatrick2016OvercomingCF} & CLIP & \val{71.71}{1.19} & \val{60.87}{1.01} & \val{70.99}{0.63} & \val{59.04}{0.85} & \val{70.43}{0.78} & \val{57.85}{1.23} & \val{70.57}{1.02} & \val{57.67}{0.47} \\
    L2P*~\cite{Wang2021LearningTP} & CLIP & \val{65.58}{0.59} & \val{53.10}{0.34} & \val{67.18}{0.78} & \val{54.92}{0.92} & \val{67.44}{0.62} & \val{54.82}{0.38} & \val{68.14}{0.55} & \val{55.61}{0.29} \\
    DualPrompt*~\cite{Wang2022DualPromptCP} & CLIP & \uuline{\val{72.50}{0.56}} & \val{62.25}{0.74} & \val{73.10}{0.85} & \uuline{\val{63.18}{0.67}} & \val{73.81}{0.77} & \uuline{\val{64.00}{0.68}} & \val{74.44}{0.46} & \uuline{\val{64.58}{0.94}} \\
    S-Prompt~\cite{Wang2022SPromptsLW} & CLIP & \val{62.28}{0.32} & \val{50.18}{0.36} & \val{67.52}{0.58} & \val{55.81}{0.46} & \val{69.85}{0.21} & \val{58.60}{0.19} & \val{70.92}{0.37} & \val{59.95}{0.28} \\
    CODA-Prompt~\cite{Smith2022CODAPromptCD} & CLIP & \val{72.43}{0.95} & \uuline{\val{62.38}{0.87}} & \uuline{\val{73.22}{0.62}} & \val{63.10}{0.76} & \uuline{\val{73.87}{0.67}} & \val{63.66}{0.59} & \uuline{\val{74.47}{0.97}} & \val{64.24}{0.27} \\
    InfLORA*~\cite{liang2024inflora} & CLIP & \val{72.39}{0.49} & \val{61.79}{0.67} & \val{72.05}{0.86} & \val{60.90}{0.77} & \val{71.83}{0.64} & \val{60.04}{0.95} & \val{71.47}{0.37} & \val{58.93}{0.65} \\
    CP-Prompt~\cite{feng2024cp} & CLIP & \val{70.02}{0.61} & \val{58.16}{0.58} & \val{71.58}{0.97} & \val{60.27}{0.67} & \val{72.52}{0.94} & \val{61.82}{0.88} & -- & -- \\
    \pgo~\cite{mukherjee2026textscpgoben} & CLIP & \valb{74.27}{0.11} & \uline{\val{63.76}{0.19}} & \valb{74.36}{0.25} & \uline{\val{63.92}{0.28}} & \valb{74.93}{0.16} & \uline{\val{64.48}{0.29}} & \valb{75.51}{0.17} & \valb{66.93}{0.26} \\
    \midrule
    CVLC & CLIP & \uline{\val{73.75}{0.08}} & \valb{63.95}{0.28} & \uline{\val{74.19}{0.08}} & \valb{65.06}{0.19} & \uline{\val{74.78}{0.06}} & \valb{65.78}{0.09} & \uline{\val{75.11}{0.15}} & \uline{\val{66.35}{0.23}} \\
    & $\Delta$ & -0.52 & +0.19 & -0.17 & +1.14 & -0.15 & +1.30 & -0.4 & -0.58 \\
    \bottomrule
\end{tabular}
}
\caption{Comparison of different methods on the DomainNet dataset. The highest values are shown in boldface, the second-highest are underlined, and the third-highest are double underlined.}
\end{table*}

\begin{figure*}[t!]
   \centering
   \includegraphics[width=0.9\linewidth]{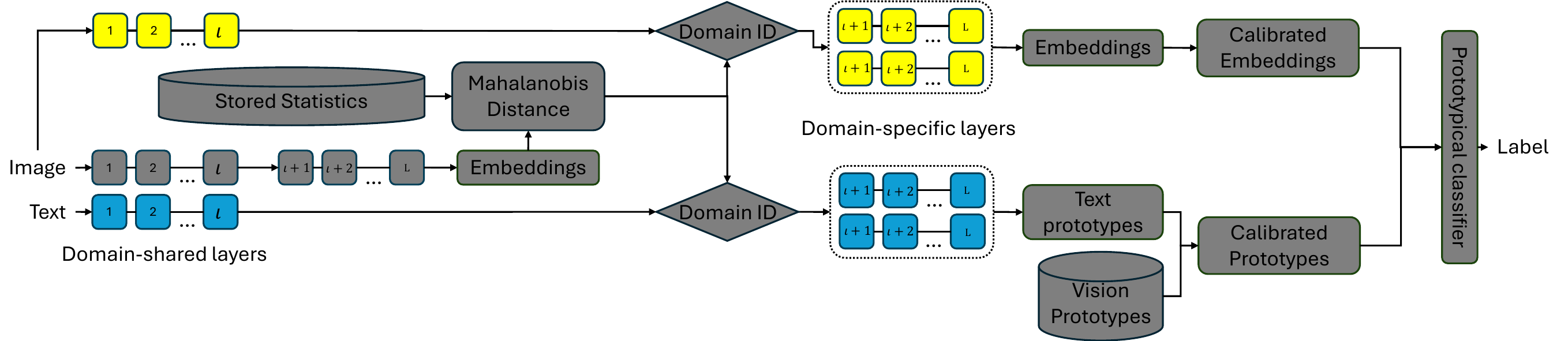}
   \caption{An overview of the CVLC at inference. The vision encoder blocks are shown in yellow, and the text encoder blocks are displayed in Blue. The embeddings from the frozen backbone, along with the stored means and covariances, are used to determine the domain ID. The output of the shared layers (with PEFT parameters) will be passed through the domain-specific layers (with PEFT parameters) to obtain the uncalibrated embeddings. After calibration, the vision and text prototypes are also being calibrated, and the maximum similarities between the calibrated embeddings and the calibrated prototypes is used to determine the labels.}
   \label{fig:arch_inference}
\end{figure*}

\subsection{Catastrophic Forgetting Mitigation}
Although CVLC implements a parameter isolation strategy with domain-specific components, the CF problem persists due to the representational drift. That is, the shared coalescent matrices are updated from $C_{s,1}^{t-1},C_{s,2}^{t-1}$ to $C_{s,1}^{t},C_{s,2}^{t}$ and compromises the validity of previous prototypes $\mu_{c}^{1:(t-1)}$. Note that CVLC utilizes domain-specific prototypes, where as a result previous prototypes may no longer be valid to cover their corresponding domains due to the representational drifts. Motivated by \cite{Zhang_2026_CVPR}, a prototype correction strategy is performed to allow old prototypes $\mu_{c}^{1:(t-1)}$ to keep pace with the representational drift issue. In particular, the old prototypes are adjusted:
\begin{equation}
    \mu_{c}^{1:(t-1)}=\mu_{c}^{1:(t-1)}+\Delta_{c}^{1:(t-1)},
\end{equation}
where $\mu_{c}^{1:(t-1)}$ denotes the prototypes of the $c-th$ class from the first domain to the previous domain, $(t-1)$ and $\Delta_{c}^{1:(t-1)}$ stands for the compensation term of the $c-th$ class from the first domain to the previous domain. The compensation is based on the fact that the domain shift can be measured by prototype displacements across two domains, e.g., $\mu_{c}^{1}$ and $\mu_{c}^{t}$. Hence, the displacement applied to the $(t-1)$ domain is defined:
\begin{equation}
    \delta_{c}^{t-1}=\mu_{c}^{t} - \mu_{c}^{t-1}
\end{equation}
\begin{equation}
    \Delta_{c}^{t-1}=\sum_{j=1}^{M}w_{c,j}\delta_{c}^{t-1}=\sum_{j=1}^{M}\frac{\exp{(\gamma.s_{c,j})}}{\sum_{l=1}^{M}\exp{(\gamma.s_{c,l})}}\delta_{c}^{t-1}
\end{equation}
where $\gamma$ is a temperature governing the sharpness of the distribution. $s_{c,j}$ stands for the cosine similarity between two prototypes $\mu_{c}^{t-1}$ and $\mu_{j}^{t-1}$ of $(t-1)$ domain while $M$ denotes the number of target classes. In other words, the prototype rectification step is performed to all previous prototypes by examining their displacements in respect to the current prototypes.

\subsection{Domain ID Predictions}
Since the domain label or oracle is not available for inferences, CVLC is equipped by a domain ID prediction strategy to infer correct domain-specific DCPs and classification heads or prototypes. A non-parametric strategy is proposed here to detect the domain IDs. It is motivated by the fact that the domain style can be described by its statistics $(\mu_{t}, V_{t})$ denoting respectively the mean and covariance of the $t-th$ domain \cite{Paeedeh2024FewShotCI}. The domain ID is simply predicted as the closest domain to the current embedding using the Mahalanobis distance $t^{*}=\arg\min_{t\in T}(z_{i}-\mu_{t})^{T}V_{t}^{-1}(z_{i}-\mu_{t})$. Our approach is simple and doesn't include any learnable parameters \cite{Paeedeh2025ContinualKC}, \cite{Wang2025BoostingDI}. Detailed algorithms and complexity analysis are offered in \textbf{the supplementary materials.} \cref{fig:arch_inference} visualizes the flow of CVLC's inference process where it first detects the domain ID to select domain-specific components, i.e., DCPs and prototypes. The prototypes are further calibrated and fused across both modalities returning the final predictions. Flows of the base and incremental training process are offered in \textbf{the supplemental document}.

\begin{table}[!htbp]
    \centering
    \resizebox{\linewidth}{!}{
    \begin{tabular}{llllll}
         Method      & Vision, shared   & Vision, specific    & Text, shared   & Text, specific      & Total \\
         \toprule
         DCP          & 65,536           & $T \times 32,768$   & 65,536         & $T \times 32,768$   & $131,072 + (T \times 65,536)$ \\
         LoRA        & 73,728           & $T \times 36,864$   & 65,536         & $T \times 32,768$   & $139,264 + (T \times 69,632) $ \\
         Prompt      & 49,152           & $T \times 24,576 $  & 32,768         & $T \times 16,384$   & $81,920 + (T \times 40,960)$ \\
    \end{tabular}
    }
    \caption{Number of parameters for the PEFT methods}
    \label{tab:num_parameters}
\end{table}
\begin{table*}[!htbp]
    \centering
    \resizebox{\textwidth}{!}{
    \begin{tabular}{l c c c c c c c c c}
    \toprule
    Method & Backbone & \multicolumn{2}{c}{1-shot} & \multicolumn{2}{c}{2-shot} & \multicolumn{2}{c}{4-shot} & \multicolumn{2}{c}{8-shot} \\
    \cmidrule(lr){3-4} \cmidrule(lr){5-6} \cmidrule(lr){7-8} \cmidrule(lr){9-10}
     & & $AA^*$(↑) & $FA^*$(↑) & $AA^*$(↑) & $FA^*$(↑) & $AA^*$(↑) & $FA^*$(↑) & $\$AA^*$(↑) & $FA^*$(↑) \\
    \midrule
    CVLC & CLIP & \valb{95.74}{0.13} & \valb{94.35}{0.07} & \valb{96.21}{0.17} & \valb{95.11}{0.20} & \val{96.44}{0.14} & \valb{95.43}{0.13} & \val{96.96}{0.06} & \valb{96.08}{0.13} \\
    \midrule
    CVLC (LoRAs) & CLIP & \val{94.91}{0.30} & \val{93.77}{0.29} & \val{95.37}{0.07} & \val{94.36}{0.09} & \val{95.47}{0.11} & \val{94.31}{0.30} & \val{95.83}{0.08} & \val{94.95}{0.11} \\
    CVLC (Prompts) & CLIP & \val{93.77}{0.17} & \val{92.25}{0.14} & \val{94.16}{0.05} & \val{92.59}{0.23} & \val{94.15}{0.06} & \val{92.73}{0.13} & \val{94.47}{0.29} & \val{93.14}{0.53} \\
    CVLC (w/o LSR) & CLIP & \val{95.69}{0.10} & \val{94.33}{0.10} & \val{96.04}{0.21} & \val{94.87}{0.20} & \valb{96.49}{0.19} & \val{95.43}{0.26} & \valb{96.99}{0.12} & \val{95.95}{0.08} \\
    CVLC (All layers with task-specific DCPs) & CLIP & \val{91.08}{0.38} & \val{87.32}{1.06} & \val{90.76}{0.13} & \val{86.39}{0.30} & \val{90.67}{0.23} & \val{86.17}{0.36} & \val{90.24}{0.12} & \val{85.29}{0.19} \\
    CVLC (w/o syn. + Single template) & CLIP & \val{95.53}{0.10} & \val{94.15}{0.24} & \val{95.99}{0.06} & \val{94.91}{0.14} & \val{96.29}{0.09} & \val{95.23}{0.06} & \val{96.75}{0.06} & \val{95.67}{0.17} \\
    \bottomrule
    \end{tabular}
    }
    \caption{Ablation studies on the CORe50 dataset}
    \label{tab:ablation_studies}
\end{table*}

\section{Experiments}
\subsection{Datasets}
We follow \cite{mukherjee2026textscpgoben}, where our algorithm, CVLC, is numerically validated with three benchmark datasets: CDDB-Hard \cite{li2023continual}, Core50 \cite{lomonaco2017core50}, and the clean version of the DomainNet \cite{peng2019moment}. These datasets possess different properties in terms of the number of classes and the number of domains. We follow the same domain orders as \cite{mukherjee2026textscpgoben} to ensure fair comparisons, where the first domain serves as the base domain while the rest are designated as the incremental domains. 

\subsection{Evaluation Metrics}
As with \cite{mukherjee2026textscpgoben}, CVLC is evaluated across 1, 2, 4, and 8 shots. We also provide our numerical results for 5 shots to enable comparisons with \cite{Li_2026_CVPR}, reporting only 5-shot results. Two standard evaluation metrics \cite{mukherjee2026textscpgoben}, namely average accuracy (AA) and forgetting evaluation (FA), are applied to measure the performance of consolidated algorithms. AA calculates the mean classification accuracy across all domains encountered so far, while FA computes the mean accuracy on a domain after the model is trained on subsequent domains. We compute the overall AA* and overall FA* by averaging the AAs and FAs across all sessions. Finally, the average and standard deviation of AA* and FA* across five random seeds are reported for CDDB and three random seeds for CORe50 and DomainNet.

\subsection{Baseline Algorithms}
We compare our algorithms comprehensively against 10 recently published algorithms: DyTox \cite{douillard2022dytox}, LwF \cite{Li2016LearningWF}, EwC \cite{Kirkpatrick2016OvercomingCF}, L2P \cite{Wang2021LearningTP}, DualPrompt \cite{Wang2022DualPromptCP}, S-Prompt \cite{Wang2022SPromptsLW}, CODA-Prompt \cite{Smith2022CODAPromptCD}, InfLORA \cite{liang2024inflora}, CP-Prompt \cite{feng2024cp}, PGO-BEn \cite{mukherjee2026textscpgoben}. Besides, an additional comparison is performed against \cite{Li_2026_CVPR} under the 5-shot regime in \textbf{the supplemental document}. 

\subsection{Implementation Details}
We utilized the ViT-B/16 variant of the CLIP for fair comparisons. To ensure the $K$-shot setting, we include only classes with at least $K + 1$ samples ($K$ for training and at least one for testing) across all domains. This restriction only affects the DomainNet dataset, where 39/345 classes were discarded, 9/39 of which have no training samples. More details are provided in \textbf{the supplementary material}. Furthermore, for the CoRE50 dataset, each class has 5 subcategories, which are not helpful. We replaced them with clearer class names, such as "can of Pepsi" or "can of Coca-Cola". Each CP in the DCPs is defined as a single matrix shared across all attention heads in all experiments.

\subsection{Numerical Results}
\cref{tab:cddb} reports numerical results of consolidated algorithms in the CDDB dataset. It is observed that CVLC outperforms other algorithms with large margins. That is, the closest gap to PGO-BEn is $4.95\%$ for AA* in the 4-shot regime, while other results exhibit significant superiority of CVLC, i.e., $12.64\%$ for AA* and $15.70\%$ for FA* in the 1-shot configuration. Numerical results of consolidated algorithms for the CORe50 dataset are presented in \cref{tab:core50}. Our findings in the CORe50 dataset confirm the advantage of CVLC against other consolidated algorithms. The largest gap exists in the 1-shot configuration, where ours beats PGO-BEn by $12.55\%$ for AA* and $16.21\%$ for FA*, whereas the smallest improvements are found in the 8-shot setting, i.e., $8.53\%$ for AA* and $10.74\%$ for FA*. Besides, our algorithm is highly competitive in the challenging DomainNet dataset, as shown in \cref{tab:domain_net}. It ranks second for AA* across all shots, but the margin is trivial relative to PGO-BEn, i.e., less than $1\%$. Our algorithm surpasses PGO-BEn for FA* in 1-shot, 2-shot, and 4-shot configurations.    

\subsection{Ablation Studies}
In this section, the contribution of each component in CVLC is examined. \cref{tab:ablation_studies} summarizes our numerical results across different configurations. \par

\subsubsection{Parameter Efficient Fine-Tuning Method}
To compare the DCPs with LoRAs, following the \cite{zhang2026exemplar}, we use LoRAs for the query and value projections. Moreover, we set the rank to 3 for vision and 4 for the text encoder, which is nearly as many elements as the DCPs use for a fair comparison. Furthermore, the prompt lengths are set to 8, as recommended in PGO-Ben. The number of parameters across DCP, LoRA, and Prompt is provided in \cref{tab:num_parameters}. It is clearly seen here that DCP produces the most encouraging performance across all shots. Using LoRA worsens numerical results by around $1\%$ while incurring fewer parameters than LoRA with DCP. On the other side, prompt deteriorates our numerical results by $2\%$ with a moderate increase of parameters in DCP compared with prompt. 

\subsubsection{Latent Space Reservation}
The absence of latent space reservation (LSR) also drops the performance of CVLC. It is perceived under the 2-shot regime that AA* declines by $0.2\%$ while FA* decreases by $0.3\%$. This module contributes to prepare the model in the base task for unseen incremental domains, i.e., it reserves the latent space by training with imaginary classes. 

\subsubsection{Synonyms and Multiple Templates}
This part offers an analysis of synonyms and multiple templates to the overall performance of CVLC. It is seen from \cref{tab:ablation_studies} that the absence of synonyms and multiple templates to detail semantics of an object leads to consistent performance drops across all shots. Although the performance gap is moderate at around $0.2-0.5\%$, it occurs across all shots confirming the advantage of our proposal. 

\subsubsection{Common and Domain-Specific Information}
We configure CVLC with only domain-specific DCPs without any shared DCPs as adopted in \cite{Wang2021LearningTP,mukherjee2026textscpgoben}. Note that CVLC accommodates both the shared and domain-specific DCPs where the first $l$ transformer layers are assigned to the shared DCP while the remainder of $L-l$ transformer blocks are dedicated to the domain-specific DCPs. Our finding reported in \cref{tab:ablation_studies} shows significant performance drops across all shots when structuring CVLC with only domain-specific DCPs. It signals the efficacy of domain-shared knowledge in addition to that of domain-specific details.  

\section{Conclusion}
This paper puts into perspective a novel algorithm, termed Continual Vision Language Consolidation (CVLC) to address challenging FSDIL problems. CVLC is constructed under the CLIP backbone whose the text and vision encoders are inserted with an alternative PEFT approach, dual coalescent projection (DCP). The concept of multi-templates and synonyms is proposed to craft the text prototype further calibrated and fused with the visual prototype. Another innovation is seen in both domain-invariant and domain-specific knowledge where the first $l$ transformer blocks apply the shared DCPs while the remaining $L-l$ layers implement the domain-specific DCPs. The training process is carried out using the idea of latent space reservation in the base domain regularizing the model to go beyond the original classes such that incremental domains can be handled seamlessly while the idea of prototype correction is integrated to deal with representational drift issues causing old prototypes to be outdated. 

Our rigorous experiments demonstrate the advantage of CVLC where it outperforms prior arts across all shots with up to a $16\%$ margin in the CDDB and CORe50 datasets. Ours also exhibits highly competitive performance compared to prior arts in the DomainNet dataset. Our ablation study further confirms tha advantage of learning components of CVLC showing their positive contributions.   

{
    \small
    \bibliographystyle{ieeenat_fullname}
    \bibliography{references}

@String(CVPR= {IEEE Conf. Comput. Vis. Pattern Recog.})

@String(ECCV= {Eur. Conf. Comput. Vis.})

@String(AAAI = {AAAI})

@String(CVPR  = {CVPR})

@String(ECCV  = {ECCV})

@book{chen2018lifelong,
  title={Lifelong machine learning},
  author={Chen, Zhiyuan and Liu, Bing},
  volume={1},
  year={2018},
  publisher={Springer}
}

@article{Parisi2018ContinualLL,
  title={Continual Lifelong Learning with Neural Networks: A Review},
  author={German Ignacio Parisi and Ronald Kemker and Jose L. Part and Christopher Kanan and Stefan Wermter},
  journal={Neural networks : the official journal of the International Neural Network Society},
  year={2018},
  volume={113},
  pages={
          54-71
        },
  url={https://api.semanticscholar.org/CorpusID:73497737}
}

@article{Masana2020ClassIncrementalLS,
  title={Class-Incremental Learning: Survey and Performance Evaluation on Image Classification},
  author={Marc Masana and Xialei Liu and Bartlomiej Twardowski and Mikel Menta and Andrew D. Bagdanov and Joost van de Weijer},
  journal={IEEE Transactions on Pattern Analysis and Machine Intelligence},
  year={2020},
  volume={45},
  pages={5513-5533},
  url={https://api.semanticscholar.org/CorpusID:234353728}
}

@article{Zhou2023ClassIncrementalLA,
  title={Class-Incremental Learning: A Survey.},
  author={Da-Wei Zhou and Qiwen Wang and Zhiyuan Qi and Han-Jia Ye and De-chuan Zhan and Ziwei Liu},
  journal={IEEE transactions on pattern analysis and machine intelligence},
  year={2023},
  volume={PP},
  url={https://api.semanticscholar.org/CorpusID:256627357}
}

@article{Kirkpatrick2016OvercomingCF,
  title={Overcoming catastrophic forgetting in neural networks},
  author={James Kirkpatrick and Razvan Pascanu and Neil C. Rabinowitz and Joel Veness and Guillaume Desjardins and Andrei A. Rusu and Kieran Milan and John Quan and Tiago Ramalho and Agnieszka Grabska-Barwinska and Demis Hassabis and Claudia Clopath and Dharshan Kumaran and Raia Hadsell},
  journal={Proceedings of the National Academy of Sciences},
  year={2016},
  volume={114},
  pages={3521 - 3526},
  url={https://api.semanticscholar.org/CorpusID:4704285}
}

@article{lopez2017gradient,
  title={Gradient episodic memory for continual learning},
  author={Lopez-Paz, David and Ranzato, Marc'Aurelio},
  journal={Advances in neural information processing systems},
  volume={30},
  year={2017},
  url={https://api.semanticscholar.org/CorpusID:37308416}
}

@article{Chaudhry2019OnTE,
  title={On Tiny Episodic Memories in Continual Learning},
  author={Arslan Chaudhry and Marcus Rohrbach and Mohamed Elhoseiny and Thalaiyasingam Ajanthan and Puneet Kumar Dokania and Philip H. S. Torr and Marc'Aurelio Ranzato},
  journal={arXiv: Learning},
  year={2019},
  doi={https://doi.org/10.48550/arXiv.1902.10486},
  url={https://api.semanticscholar.org/CorpusID:173188188}
}

@article{Zenke2017ContinualLT,
  title={Continual Learning Through Synaptic Intelligence},
  author={Friedemann Zenke and Ben Poole and Surya Ganguli},
  journal={Proceedings of machine learning research},
  year={2017},
  volume={70},
  pages={
          3987-3995
        },
  url={https://api.semanticscholar.org/CorpusID:10409742}
}

@article{aljundi2018memory,
  title={Memory aware synapses: Learning what (not) to forget},
  author={Aljundi, Rahaf and Babiloni, Francesca and Elhoseiny, Mohamed and Rohrbach, Marcus and Tuytelaars, Tinne},
  booktitle={Proceedings of the European conference on computer vision (ECCV)},
  pages={139--154},
  year={2018},
  doi={https://doi.org/10.48550/arXiv.1711.09601},
  url={https://api.semanticscholar.org/CorpusID:4254748}
}

@article{Li2016LearningWF,
  title={Learning without Forgetting},
  author={Zhizhong Li and Derek Hoiem},
  journal={IEEE Transactions on Pattern Analysis and Machine Intelligence},
  year={2016},
  volume={40},
  pages={2935-2947},
  url={https://api.semanticscholar.org/CorpusID:4853851}
}

@article{Schwarz2018ProgressC,
  title={Progress \& Compress: A scalable framework for continual learning},
  author={Jonathan Schwarz and Wojciech M. Czarnecki and Jelena Luketina and Agnieszka Grabska-Barwinska and Yee Whye Teh and Razvan Pascanu and Raia Hadsell},
  journal={ArXiv},
  year={2018},
  volume={abs/1805.06370},
  doi={https://doi.org/10.48550/arXiv.1711.09601},
  url={https://api.semanticscholar.org/CorpusID:21718339}
}

@inproceedings{Paik2019OvercomingCF,
  title={Overcoming Catastrophic Forgetting by Neuron-level Plasticity Control},
  author={Inyoung Paik and Sangjun Oh and Taeyeong Kwak and Injung Kim},
  booktitle={AAAI Conference on Artificial Intelligence},
  year={2019},
  doi={https://doi.org/10.48550/arXiv.1907.13322},
  url={https://api.semanticscholar.org/CorpusID:199001153}
}

@article{Mao2021ContinualLV,
  title={Continual learning via inter-task synaptic mapping},
  author={Fubing Mao and Weiwei Weng and Mahardhika Pratama and Edward Kien Yee Yapp},
  journal={ArXiv},
  year={2021},
  volume={abs/2106.13954},
  doi={https://doi.org/10.1016/j.knosys.2021.106947},
  url={https://api.semanticscholar.org/CorpusID:233710658}
}

@article{Cha2020CPRCR,
  title={CPR: Classifier-Projection Regularization for Continual Learning},
  author={Sungmin Cha and Hsiang Hsu and Fl{\'a}vio du Pin Calmon and Taesup Moon},
  journal={ArXiv},
  year={2020},
  volume={abs/2006.07326},
  doi={https://doi.org/10.48550/arXiv.2006.07326},
  url={https://api.semanticscholar.org/CorpusID:219636462}
}

@article{Yoon2017LifelongLW,
  title={Lifelong Learning with Dynamically Expandable Networks},
  author={Jaehong Yoon and Eunho Yang and Jeongtae Lee and Sung Ju Hwang},
  journal={ArXiv},
  year={2017},
  volume={abs/1708.01547},
  doi={https://doi.org/10.48550/arXiv.1708.01547},
  url={https://api.semanticscholar.org/CorpusID:3693512}
}

@article{Li2019LearnTG,
  title={Learn to Grow: A Continual Structure Learning Framework for Overcoming Catastrophic Forgetting},
  author={Xilai Li and Yingbo Zhou and Tianfu Wu and Richard Socher and Caiming Xiong},
  journal={ArXiv},
  year={2019},
  volume={abs/1904.00310},
  doi={https://doi.org/10.48550/arXiv.1904.00310},
  url={https://api.semanticscholar.org/CorpusID:90259576}
}

@article{Xu2021AdaptivePC,
  title={Adaptive Progressive Continual Learning},
  author={Ju Xu and Jin Ma and Xuesong Gao and Zhanxing Zhu},
  journal={IEEE Transactions on Pattern Analysis and Machine Intelligence},
  year={2021},
  volume={44},
  pages={6715-6728},
  url={https://api.semanticscholar.org/CorpusID:235767757}
}

@inproceedings{Pratama2021UnsupervisedCL,
  title={Unsupervised Continual Learning via Self-Adaptive Deep Clustering Approach},
  author={Mahardhika Pratama and Andri Ashfahani and Edwin David Lughofer},
  booktitle={CSSL},
  year={2021},
  doi={https://doi.org/10.48550/arXiv.2106.14563},
  url={https://api.semanticscholar.org/CorpusID:235658071}
}

@article{Rebuffi2016iCaRLIC,
  title={iCaRL: Incremental Classifier and Representation Learning},
  author={Sylvestre-Alvise Rebuffi and Alexander Kolesnikov and G. Sperl and Christoph H. Lampert},
  journal={2017 IEEE Conference on Computer Vision and Pattern Recognition (CVPR)},
  year={2016},
  pages={5533-5542},
  doi={https://doi.org/10.48550/arXiv.1611.07725},
  url={https://api.semanticscholar.org/CorpusID:206596260}
}

@article{Chaudhry2018EfficientLL,
  title={Efficient Lifelong Learning with A-GEM},
  author={Arslan Chaudhry and Marc'Aurelio Ranzato and Marcus Rohrbach and Mohamed Elhoseiny},
  journal={ArXiv},
  year={2018},
  volume={abs/1812.00420},
  doi={https://doi.org/10.48550/arXiv.1812.00420},
  url={https://api.semanticscholar.org/CorpusID:54443381}
}

@inproceedings{Chaudhry2019UsingHT,
  title={Using Hindsight to Anchor Past Knowledge in Continual Learning},
  author={Arslan Chaudhry and Albert Gordo and Puneet Kumar Dokania and Philip H. S. Torr and David Lopez-Paz},
  booktitle={AAAI Conference on Artificial Intelligence},
  year={2021},
  doi={https://doi.org/10.48550/arXiv.2002.08165},
  url={https://api.semanticscholar.org/CorpusID:210957697}
}

@article{Buzzega2020DarkEF,
  title={Dark Experience for General Continual Learning: a Strong, Simple Baseline},
  author={Pietro Buzzega and Matteo Boschini and Angelo Porrello and Davide Abati and Simone Calderara},
  journal={ArXiv},
  year={2020},
  volume={abs/2004.07211},
  doi={https://doi.org/10.48550/arXiv.2004.07211},
  url={https://api.semanticscholar.org/CorpusID:215768806}
}

@inproceedings{Shin2017ContinualLW,
  title={Continual Learning with Deep Generative Replay},
  author={Hanul Shin and Jung Kwon Lee and Jaehong Kim and Jiwon Kim},
  booktitle={Neural Information Processing Systems},
  year={2017},
  doi={https://doi.org/10.48550/arXiv.1705.08690},
  url={https://api.semanticscholar.org/CorpusID:1888776}
}

@article{Rakaraddi2022ReinforcedCL,
  title={Reinforced Continual Learning for Graphs},
  author={Appan Rakaraddi and Siew-Kei Lam and Mahardhika Pratama and Marcus Vin{\'i}cius de Carvalho},
  journal={Proceedings of the 31st ACM International Conference on Information \& Knowledge Management},
  year={2022},
          doi={https://doi.org/10.48550/arXiv.2209.01556},
  url={https://api.semanticscholar.org/CorpusID:252089650}
}

@inproceedings{Dam2022ScalableAO,
  title={Scalable Adversarial Online Continual Learning},
  author={T. Dam and Mahardhika Pratama and Md Meftahul Ferdaus and Sreenatha G. Anavatti and Hussein Abbas},
  booktitle={ECML/PKDD},
  year={2022},
          doi={https://doi.org/10.48550/arXiv.2209.01558},
  url={https://api.semanticscholar.org/CorpusID:252089827}
}

@inproceedings{VinciusdeCarvalho2022ClassIncrementalLV,
  title={Class-Incremental Learning via Knowledge Amalgamation},
  author={Marcus Vin{\'i}cius de Carvalho and Mahardhika Pratama and Jie Zhang and Yajuan San},
  booktitle={ECML/PKDD},
  year={2022},
doi={https://doi.org/10.48550/arXiv.2209.02112},
  url={https://api.semanticscholar.org/CorpusID:252090326}
}

@article{Ashfahani2021UnsupervisedCL,
  title={Unsupervised Continual Learning in Streaming Environments},
  author={Andri Ashfahani and Mahardhika Pratama},
  journal={IEEE Transactions on Neural Networks and Learning Systems},
  year={2021},
  volume={34},
  pages={9992-10003},
  url={https://api.semanticscholar.org/CorpusID:237572299}
}

@article{Masum2023AssessorGuidedLF,
  title={Assessor-Guided Learning for Continual Environments},
  author={M. A. Ma'sum and Mahardhika Pratama and Edwin David Lughofer and Weiping Ding and Wisnu Jatmiko},
  journal={Inf. Sci.},
  year={2023},
  volume={640},
  pages={119088},
doi={https://doi.org/10.48550/arXiv.2303.11624},
  url={https://api.semanticscholar.org/CorpusID:257636622}
}

@article{Wang2021LearningTP,
  title={Learning to Prompt for Continual Learning},
  author={Zifeng Wang and Zizhao Zhang and Chen-Yu Lee and Han Zhang and Ruoxi Sun and Xiaoqi Ren and Guolong Su and Vincent Perot and Jennifer G. Dy and Tomas Pfister},
  journal={2022 IEEE/CVF Conference on Computer Vision and Pattern Recognition (CVPR)},
  pages={139--149},
  year={2022},
  doi={https://doi.org/10.48550/arXiv.2112.08654},
  url={https://api.semanticscholar.org/CorpusID:245218925}
}

@article{Wang2022DualPromptCP,
  title={DualPrompt: Complementary Prompting for Rehearsal-free Continual Learning},
  author={Zifeng Wang and Zizhao Zhang and Sayna Ebrahimi and Ruoxi Sun and Han Zhang and Chen-Yu Lee and Xiaoqi Ren and Guolong Su and Vincent Perot and Jennifer G. Dy and Tomas Pfister},
  journal={ArXiv},
  year={2022}, 
 doi={https://doi.org/10.1007/978-3-031-19809-0_36},
 pages = {631–648},
  url={https://api.semanticscholar.org/CorpusID:248085201}
}

@article{He2025CLLoRACL,
  title={CL-LoRA: Continual Low-Rank Adaptation for Rehearsal-Free Class-Incremental Learning},
  author={Jiangpeng He and Zhihao Duan and Fengqing Maggie Zhu},
  journal={ArXiv},
  year={2025},
  volume={abs/2505.24816},
  url={https://api.semanticscholar.org/CorpusID:279070575}
}

@article{Liu2025LoRASF,
  title={LoRA Subtraction for Drift-Resistant Space in Exemplar-Free Continual Learning},
  author={Xuan Liu and Xiaobin Chang},
  journal={ArXiv},
  year={2025},
  volume={abs/2503.18985},
  url={https://api.semanticscholar.org/CorpusID:277313510}
}

@article{Fukuda2024AdapterMW,
  title={Adapter Merging with Centroid Prototype Mapping for Scalable Class-Incremental Learning},
  author={Takuma Fukuda and Hiroshi Kera and Kazuhiko Kawamoto},
  journal={ArXiv},
  year={2024},
  volume={abs/2412.18219},
  url={https://api.semanticscholar.org/CorpusID:274992202}
}

@inproceedings{Zhou2024ContinualLW,
  title={Continual Learning with Pre-Trained Models: A Survey},
  author={Da-Wei Zhou and Hai-Long Sun and Jingyi Ning and Han-Jia Ye and De-chuan Zhan},
  booktitle={International Joint Conference on Artificial Intelligence},
  year={2024},
  url={https://api.semanticscholar.org/CorpusID:267312447}
}

@article{vandeVen2019ThreeSF,
  title={Three scenarios for continual learning},
  author={Gido M. van de Ven and Andreas Savas Tolias},
  journal={ArXiv},
  year={2019},
  volume={abs/1904.07734},
  url={https://api.semanticscholar.org/CorpusID:119309522}
}

@article{Liu2024CompositionalPF,
  title={Compositional Prompting for Anti-Forgetting in Domain Incremental Learning},
  author={Zichen Liu and Yuxin Peng and Jiahuan Zhou},
  journal={Int. J. Comput. Vis.},
  year={2024},
  volume={132},
  pages={5783-5800},
  url={https://api.semanticscholar.org/CorpusID:270778693}
}

@inproceedings{Wang2025DualCPRD,
  title={DualCP: Rehearsal-Free Domain-Incremental Learning via Dual-Level Concept Prototype},
  author={Qiang Wang and Yuhang He and Songlin Dong and Xiang Song and Jizhou Han and Haoyu Luo and Yihong Gong},
  booktitle={AAAI Conference on Artificial Intelligence},
  year={2025},
  url={https://api.semanticscholar.org/CorpusID:277272564}
}

@article{Wang2025BoostingDI,
  title={Boosting Domain Incremental Learning: Selecting the Optimal Parameters is All You Need},
  author={Qiang Wang and Xiang Song and Yuhang He and Jizhou Han and Chenhao Ding and Xinyuan Gao and Yihong Gong},
  journal={ArXiv},
  year={2025},
  volume={abs/2505.23744},
  url={https://api.semanticscholar.org/CorpusID:278996449}
}

@article{Zhou2024DualCF,
  title={Dual Consolidation for Pre-Trained Model-Based Domain-Incremental Learning},
  author={Da-Wei Zhou and Zi-Wen Cai and Han-Jia Ye and Lijun Zhang and De-Chuan Zhan},
  journal={ArXiv},
  year={2024},
  volume={abs/2410.00911},
  url={https://api.semanticscholar.org/CorpusID:273022628}
}

@article{Paeedeh2024FewShotCI,
  title={Few-Shot Class Incremental Learning via Robust Transformer Approach},
  author={Naeem Paeedeh and Mahardhika Pratama and Sunu Wibirama and Wolfgang Mayer and Zehong Cao and Ryszard Kowalczyk},
  journal={Inf. Sci.},
  year={2024},
  volume={675},
  pages={120751},
  url={https://api.semanticscholar.org/CorpusID:269741067}
}

@inproceedings{peng2019moment,
  title={Moment matching for multi-source domain adaptation},
  author={Peng, Xingchao and Bai, Qinxun and Xia, Xide and Huang, Zijun and Saenko, Kate and Wang, Bo},
  booktitle={Proceedings of the IEEE/CVF international conference on computer vision},
  pages={1406--1415},
  year={2019}
}

@inproceedings{lomonaco2017core50,
  title={Core50: a new dataset and benchmark for continuous object recognition},
  author={Lomonaco, Vincenzo and Maltoni, Davide},
  booktitle={Conference on robot learning},
  pages={17--26},
  year={2017},
  organization={PMLR}
}

@inproceedings{li2023continual,
  title={A continual deepfake detection benchmark: Dataset, methods, and essentials},
  author={Li, Chuqiao and Huang, Zhiwu and Paudel, Danda Pani and Wang, Yabin and Shahbazi, Mohamad and Hong, Xiaopeng and Van Gool, Luc},
  booktitle={Proceedings of the IEEE/CVF winter conference on applications of computer vision},
  pages={1339--1349},
  year={2023}
}

@article{Smith2022CODAPromptCD,
  title={CODA-Prompt: COntinual Decomposed Attention-Based Prompting for Rehearsal-Free Continual Learning},
  author={James Smith and Leonid Karlinsky and Vyshnavi Gutta and Paola Cascante-Bonilla and Donghyun Kim and Assaf Arbelle and Rameswar Panda and Rog{\'e}rio Schmidt Feris and Zsolt Kira},
  journal={2023 IEEE/CVF Conference on Computer Vision and Pattern Recognition (CVPR)},
  year={2022},
  pages={11909-11919},
  url={https://api.semanticscholar.org/CorpusID:253801593}
}

@article{Wang2022SPromptsLW,
  title={S-Prompts Learning with Pre-trained Transformers: An Occam's Razor for Domain Incremental Learning},
  author={Yabin Wang and Zhiwu Huang and Xiaopeng Hong},
  journal={ArXiv},
  year={2022},
  volume={abs/2207.12819},
  url={https://api.semanticscholar.org/CorpusID:251066766}
}

@article{Paeedeh2025ContinualKC,
  title={Continual Knowledge Consolidation LORA for Domain Incremental Learning},
  author={Naeem Paeedeh and Mahardhika Pratama and Weiping Ding and Jimmy Cao and Wolfgang Mayer and Ryszard Kowalczyk},
  journal={ArXiv},
  year={2025},
  volume={abs/2510.16077},
  url={https://api.semanticscholar.org/CorpusID:282209751}
}

@article{
mukherjee2026textscpgoben,
title={{PGO-BEN}: Proxy-Guided Orthogonalization and Beta En-sembling for Few-Shot Domain-Incremental Learning},
author={Samrat Mukherjee and Thivyanth Venkateswaran and Eric Nuertey Coleman and Luigi Quarantiello and Julio Hurtado and Vincenzo Lomonaco and Gemma Roig and Subhasis Chaudhuri and Biplab Banerjee},
journal={Transactions on Machine Learning Research},
issn={2835-8856},
year={2026},
url={https://openreview.net/forum?id=jlb27FbHLv},
note={}
}

@article{Radford2021LearningTV,
  title={Learning Transferable Visual Models From Natural Language Supervision},
  author={Alec Radford and Jong Wook Kim and Chris Hallacy and Aditya Ramesh and Gabriel Goh and Sandhini Agarwal and Girish Sastry and Amanda Askell and Pamela Mishkin and Jack Clark and Gretchen Krueger and Ilya Sutskever},
  journal={ArXiv},
  year={2021},
  volume={abs/2103.00020},
  url={https://api.semanticscholar.org/CorpusID:231591445}
}

@inproceedings{Snell2017PrototypicalNF,
  title={Prototypical Networks for Few-shot Learning},
  author={Jake Snell and Kevin Swersky and Richard S. Zemel},
  booktitle={Neural Information Processing Systems},
  year={2017},
  url={https://api.semanticscholar.org/CorpusID:309759}
}

@article{Chen2025EnhancingFC,
  title={Enhancing Few-Shot Class-Incremental Learning via Training-Free Bi-Level Modality Calibration},
  author={Yiyang Chen and Tianyu Ding and Lei Wang and Jing Huo and Yang Gao and Wenbin Li},
  journal={2025 IEEE/CVF Conference on Computer Vision and Pattern Recognition (CVPR)},
  year={2025},
  pages={9881-9890},
  url={https://api.semanticscholar.org/CorpusID:280016402}
}

@inproceedings{verma2019manifold,
  title={Manifold mixup: Better representations by interpolating hidden states},
  author={Verma, Vikas and Lamb, Alex and Beckham, Christopher and Najafi, Amir and Mitliagkas, Ioannis and Lopez-Paz, David and Bengio, Yoshua},
  booktitle={International conference on machine learning},
  pages={6438--6447},
  year={2019},
  organization={PMLR}
}

@inproceedings{chen2025pseudo,
  title={Pseudo Informative Episode Construction for Few-Shot Class-Incremental Learning},
  author={Chen, Chaofan and Yang, Xiaoshan and Xu, Changsheng},
  booktitle={Proceedings of the AAAI Conference on Artificial Intelligence},
  volume={39},
  number={15},
  pages={15749--15757},
  year={2025}
}

@inproceedings{douillard2022dytox,
  title={Dytox: Transformers for continual learning with dynamic token expansion},
  author={Douillard, Arthur and Ram{\'e}, Alexandre and Couairon, Guillaume and Cord, Matthieu},
  booktitle={Proceedings of the IEEE/CVF conference on computer vision and pattern recognition},
  pages={9285--9295},
  year={2022}
}

@inproceedings{liang2024inflora,
  title={Inflora: Interference-free low-rank adaptation for continual learning},
  author={Liang, Yan-Shuo and Li, Wu-Jun},
  booktitle={Proceedings of the IEEE/CVF Conference on Computer Vision and Pattern Recognition},
  pages={23638--23647},
  year={2024}
}

@inproceedings{feng2024cp,
  title={CP-prompt: Composition-based cross-modal prompting for domain-incremental continual learning},
  author={Feng, Yu and Tian, Zhen and Zhu, Yifan and Han, Zongfu and Luo, Haoran and Zhang, Guangwei and Song, Meina},
  booktitle={Proceedings of the 32nd ACM International Conference on Multimedia},
  pages={2729--2738},
  year={2024}
}

@InProceedings{Li_2026_CVPR,
    author    = {Li, Yan and Shi, Yuzhu and Zhou, Kan and Zhang, Shu and He, Diqi and Zhang, Dingwen and Han, Junwei},
    title     = {Few-Shot Hybrid Incremental Learning:Continually Learning under Data Scarcity and Task Uncertainty},
    booktitle = {Proceedings of the IEEE/CVF Conference on Computer Vision and Pattern Recognition (CVPR)},
    month     = {June},
    year      = {2026},
    pages     = {32334-32344}
}

@InProceedings{Zhang_2026_CVPR,
    author    = {Zhang, Xin and Bai, Liang and Wang, Guanchao and Yang, Xian},
    title     = {Exemplar-Free Class Incremental Learning via Preserving Class-Discriminative Structure},
    booktitle = {Proceedings of the IEEE/CVF Conference on Computer Vision and Pattern Recognition (CVPR)},
    month     = {June},
    year      = {2026},
    pages     = {17979-17988}
}

@inproceedings{zhang2026exemplar,
  title={Exemplar-Free Class Incremental Learning via Preserving Class-Discriminative Structure},
  author={Zhang, Xin and Bai, Liang and Wang, Guanchao and Yang, Xian},
  booktitle={Proceedings of the IEEE/CVF Conference on Computer Vision and Pattern Recognition},
  pages={17979--17988},
  year={2026}
}

@inproceedings{liu2024few,
  title={Few-shot class incremental learning with attention-aware self-adaptive prompt},
  author={Liu, Chenxi and Wang, Zhenyi and Xiong, Tianyi and Chen, Ruibo and Wu, Yihan and Guo, Junfeng and Huang, Heng},
  booktitle={European conference on computer vision},
  pages={1--18},
  year={2024},
  organization={Springer}
}

@inproceedings{wang2024approximation,
  title={On the approximation risk of few-shot class-incremental learning},
  author={Wang, Xuan and Ji, Zhong and Liu, Xiyao and Pang, Yanwei and Han, Jungong},
  booktitle={European Conference on Computer Vision},
  pages={162--178},
  year={2024},
  organization={Springer}
}

@inproceedings{liu2025sec,
  title={Sec-prompt: Semantic complementary prompting for few-shot class-incremental learning},
  author={Liu, Ye and Yang, Meng},
  booktitle={Proceedings of the Computer Vision and Pattern Recognition Conference},
  pages={25643--25656},
  year={2025}
}

@inproceedings{park2024versatile,
  title={Versatile incremental learning: Towards class and domain-agnostic incremental learning},
  author={Park, Min-Yeong and Lee, Jae-Ho and Park, Gyeong-Moon},
  booktitle={European Conference on Computer Vision},
  pages={271--288},
  year={2024},
  organization={Springer}
}

@article{paeedeh2025cross,
  title={Cross-Domain Few-Shot Learning with Coalescent Projections and Latent Space Reservation},
  author={Paeedeh, Naeem and Pratama, Mahardhika and Kamal, Imam Mustafa and Mayer, Wolfgang and Cao, Jimmy and Kowlczyk, Ryszard},
  journal={arXiv preprint arXiv:2507.15243},
  year={2025}
}

@article{loshchilov2017decoupled,
  title={Decoupled weight decay regularization},
  author={Loshchilov, Ilya and Hutter, Frank},
  journal={arXiv preprint arXiv:1711.05101},
  year={2017}
}

@software{openai2026gpt54,
  author       = {{OpenAI}},
  title        = {GPT-5.4},
  year         = {2026},
  url          = {https://chatgpt.com},
  version      = {5.4},
  note         = {Large language model}
}

@software{google2026gemini31,
  author       = {{Google}},
  title        = {Gemini 3.1 Pro},
  year         = {2026},
  url          = {https://gemini.google.com},
  version      = {3.1 Pro}, 
  note         = {Large language model}
}
}

\clearpage
\appendix
\renewcommand{\thesection}{S\arabic{section}}
\setcounter{section}{0}
\renewcommand{\thepage}{S\arabic{page}}
\setcounter{page}{1}
\twocolumn[
  \begin{center}
    {\LARGE\bfseries Supplementary Material}
  \end{center}
]

\section{Related Works}
Continual learning (CL) is a growing research area where the goal is to address dynamic and evolving learning environments. The underlying bottleneck lies in the stability-plasticity dilemma where a plastic model is capable of adapting itself to new conditions but loses its previous knowledge due to the CF problem whereas a stable model preserves old knowledge but fails to adapt to new conditions. The CL problem consists of three sub-problems: task-incremental learning (TIL), class-incremental learning (CIL) and domain-incremental learning (DIL) \cite{vandeVen2019ThreeSF}. The TIL and the CIL are identical where the exception exists in the presence of the task IDs during the inference for the TIL. As a result, the CIL is more challenging than the TIL.

\subsection{Class-Incremental Learning}
CIL is a CL sub-problem that aims to address changing target classes. That is, every task introduces a unique set of target classes. A model is tasked to recognize new classes without forgetting previously seen classes. The key to solving the CIL problem lies in how to combat the CF problem. There are three approaches to overcome the CF problem: regularization-based approach \cite{Kirkpatrick2016OvercomingCF,Mao2021ContinualLV,Paik2019OvercomingCF,Zenke2017ContinualLT,aljundi2018memory,Li2019LearnTG,Schwarz2018ProgressC,Cha2020CPRCR}, architecture-based approach \cite{Xu2021AdaptivePC,Yoon2017LifelongLW,Ashfahani2021UnsupervisedCL,Pratama2021UnsupervisedCL,Li2019LearnTG,Rakaraddi2022ReinforcedCL}, and memory-based approach \cite{Chaudhry2018EfficientLL,Rebuffi2016iCaRLIC,Chaudhry2019OnTE,Chaudhry2019UsingHT,Buzzega2020DarkEF,Dam2022ScalableAO,VinciusdeCarvalho2022ClassIncrementalLV,Masum2023AssessorGuidedLF,lopez2017gradient,Shin2017ContinualLW}. The regularization-based approach appends a penalty term in the loss function. Although it is simple, the regularization-based approach doesn't scale well for large-scale problems because it is hard to find an overlapping region of all tasks. The architecture-based approach relies on the expand-and-freeze approach. New components are evolved to accommodate new conditions while isolating old components to prevent the CF problems. The architecture-based approach calls for the presence of task IDs which cannot be served in the CIL context. The memory-based approach is deemed the strongest approach where it applies experience replay mechanisms using an episodic memory. The use of memory imposes privacy and storage concerns. Such issue has led to the development of rehearsal-free approaches benefiting from strong generalization powers of foundation models \cite{Zhou2024ContinualLW}. The backbone network can be frozen to avoid the CF problem while adapting to new tasks via parameter-efficient fine-tuning (PEFT) methods: prompts \cite{Wang2021LearningTP,Wang2022DualPromptCP}, LoRAs \cite{He2025CLLoRACL,Liu2025LoRASF}, adapter \cite{Fukuda2024AdapterMW}. Although there exist many works to tackle the CIL problem, the DIL problem remains an open problem. 

\subsection{Domain-Incremental Learning}
DIL constitutes a CL sub-branch focusing on the domain shift problems. That is, a model is trained to be robust against varying domains. \cite{Liu2024CompositionalPF} presents the idea of a compositional prompt where each domain is assigned a prompt pool. \cite{Wang2025DualCPRD} features coarse-grained prototypes and fine-grained prototypes. A dual-level consolidation in the feature level and classifier level is proposed in \cite{Zhou2024DualCF}. \cite{Paeedeh2025ContinualKC} and \cite{Wang2025BoostingDI} share a commonality where a domain ID predictor is put forward. \cite{Paeedeh2025ContinualKC}, however, features a salient network structure where a stacked architecture of shared LoRAs and domain-specific LoRAs is proposed. Although the DIL topic has started to gain traction, these approaches assume the existence of plentiful samples per domain.

To date, the issue of data scarcity in the context of DIL remains an open research problem. That is, \cite{mukherjee2026textscpgoben} is the first and only approach to the few-shot domain-incremental learning (FSDIL) problem. The FSDIL problem extends the DIL problem, in which each incremental domain, except the base domain, carries very few samples formulated in the N-way K-shot setting. This paper addresses the FSDIL problem and distinguishes itself from \cite{mukherjee2026textscpgoben}, ignoring common information across domains. \cite{mukherjee2026textscpgoben} still relies on the original CLIP template, which is ambiguous and insufficient to describe an object. Although the prompt tuning approach \cite{mukherjee2026textscpgoben} limits the number of trainable parameters, it often suffers from inaccurate prompt selection and growing complexities. We propose an alternative of prompt tuning \cite{Wang2021LearningTP}, namely dual coalescent projection (DCP). That is, it relies on a dual learnable matrix combining the query, key and value into a single concept.

\section{Dual Coalescent Projection (DCP)}

Coalescent projection (CP) coalesces two concepts into a single concept by being used between two vector of concepts. From another perspective, it starts by noisy identity mapping of a vector, but it can change the order of the tokens, change the focus of attention when it is required, for example, in domain shifts, and it is anisotropic, as it uses Mahalanobis distance metric in the neighborhood around a vector by being able to stretch or shrink each axis, which makes it Hyper-Ellipsoidal instead of Hyper-spherical \cite{paeedeh2025cross}. DCP calculations are shown in the \cref{fig:DCP}. \par

While a standard CP between $Q$ and $K$ calibrates the attention map and modifies the routing, the projected value vectors are neglected. DCP extends the CP to the value projection, helping the network adapt content aggregation in the value projection to extract more useful information from each domain. Therefore, DCP helps a transformer learn both where to look and what to extract while remaining extremely parameter-efficient.

\begin{figure*}[htbp]
   \centering
   \includegraphics[width=\linewidth]{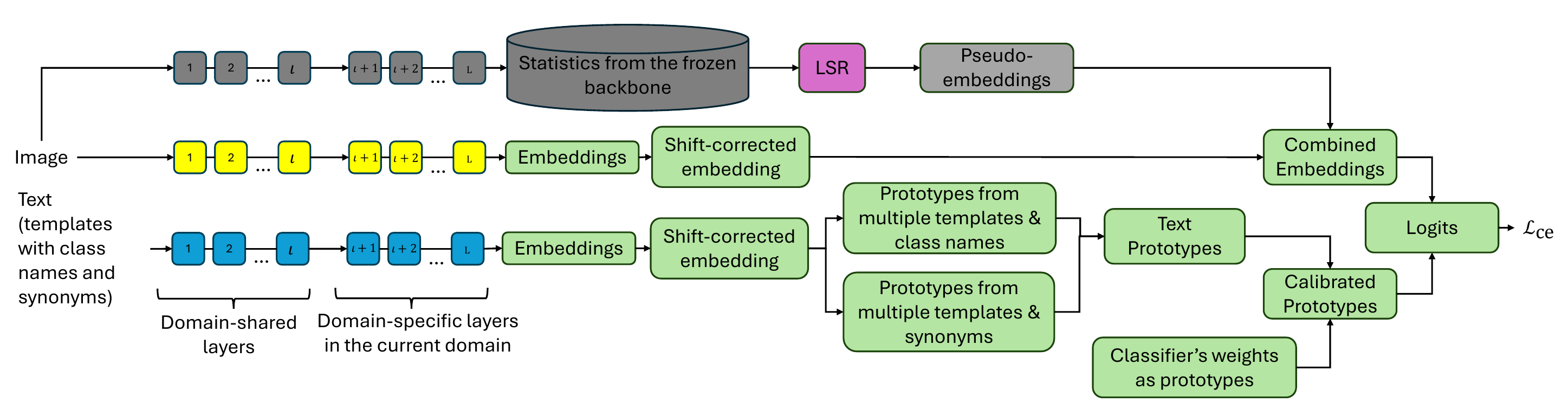}
   \caption{An overview of the CVLC in the base domain training. Frozen components are shown in gray, and differentiable intermediate calculations and components are displayed in Green.}
   \label{fig:arch_first_task}
\end{figure*}

\begin{figure*}[htbp]
   \centering
   \includegraphics[width=\linewidth]{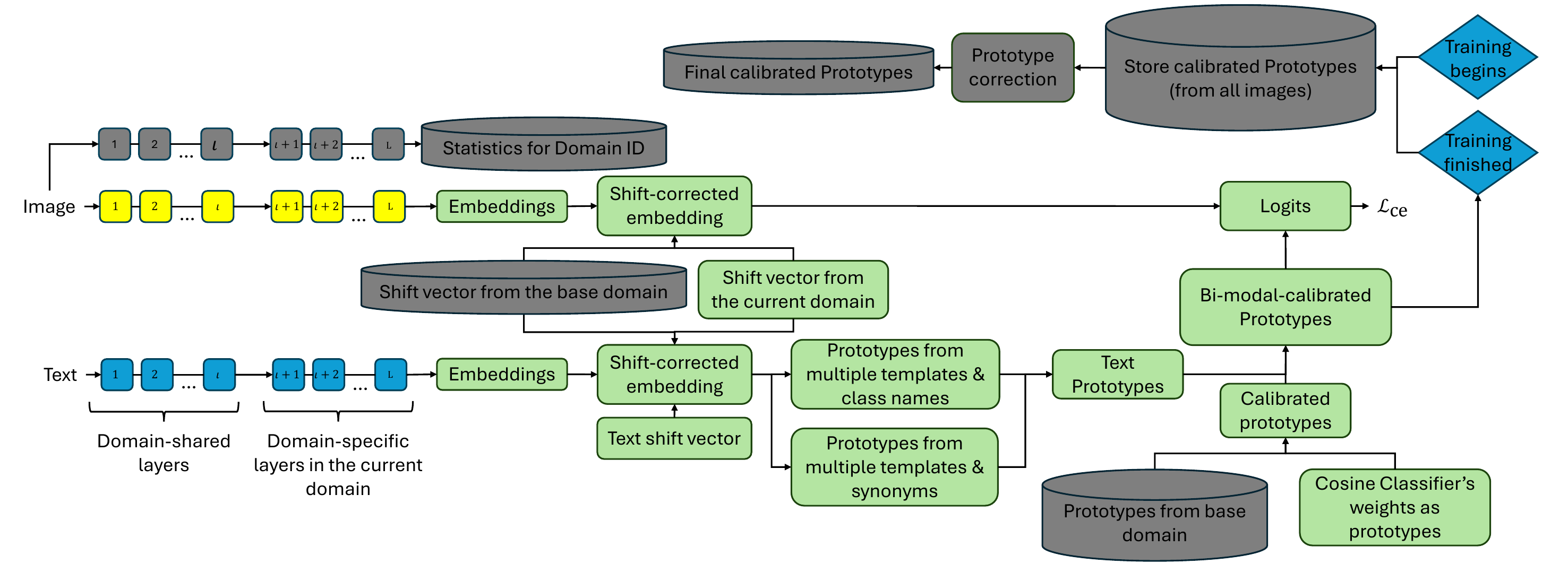}
   \caption{An overview of the CVLC in the incremental tasks. Frozen components are shown in gray, and differentiable intermediate calculations and components are displayed in Green.}
   \label{fig:arch_inc_tasks}
\end{figure*}

\section{Algorithm}

\cref{fig:arch_first_task} and \cref{fig:arch_inc_tasks} show the overview of the CVLC architecture and process in the base and incremental task training phases. In all domains, the statistics (means and covariances) are calculated from all training samples using the frozen vision encoder for domain ID detection.  In the base task, the statistics from the frozen backbone are also used to generate pseudo-embeddings with the LSR, which are then combined with the vision embeddings before calculating the logits and cross-entropy loss. In training, we use the shared DCPs and each domain's specific DCPs to obtain the embeddings. Next, learnable shift vectors correct the shifts. For the text encoder, we compute prototypes for real class names and their synonyms to obtain the final text prototypes. Finally, the classifier's weights and, in the incremental tasks, the base domain's prototypes are also used to calibrate the text prototypes. Finally, the bi-modal-calibrated prototypes are utilized to obtain the logits and compute the CE loss. For prototype correction, we compute calibrated prototypes from all training samples at the beginning and end of training in each domain. It is worth noting that in the incremental tasks, the learned domain-specific parameters are copied from the base domain, where they were well-trained on many samples. \par

The pseudo-code of the CVLC is displayed in \cref{alg:pseudo-code} and \cref{alg:LSR}. C1-C7 are learnable coefficients, the shift matrix includes the learnable shift vectors for each domain, and shift\_text\_vector is a single learnable vector for the text embeddings.

\begin{algorithm}[htbp]
  \caption{PyTorch style pseudo-code for the CVLC}
  \label{alg:pseudo-code}
  \definecolor{codeblue}{rgb}{0.25,0.5,0.5}
  \lstset{
      basicstyle=\fontsize{7.2pt}{7.2pt}\ttfamily\bfseries,
      commentstyle=\fontsize{7.2pt}{7.2pt}\color{codeblue},
      keywordstyle=\fontsize{7.2pt}{7.2pt},
      language=python,
      breaklines=true,
      frame=single,
      aboveskip=0pt,
      belowskip=0pt,
    }
  \begin{lstlisting}[language=python]

def train_all_domains():
    for domain_id in range(total_domains):
        train_one_domain()
        evaluate()

def train_one_domain():
    if domain_id == 0:
        stats_for_LSR = compute_stats_from_frozen_backbone()
        train(stats_for_LSR)
    else:
        prototypes_before_training = compute_prototypes()
        train()
        prototypes_after_training = compute_prototypes()
        correct_prototypes( # Eqs. 24-26
            prototypes_before_training, prototypes_after_training)

def train(stats_for_LSR=None)
    for i in range(epochs):
        train_one_epoch(stats_for_LSR)

def train_one_epoch(stats_for_LSR):
    for images, labels in train_loader:
        logits, labels = forward_multimodal(images, labels, stats_for_LSR)
        loss = CE_loss(logits, label)
        optimize(loss)

def forward_multimodal(images, labels, stats_for_LSR):
    embeddings_vis = forward_vision(images)
    if domain_id == 0:  # LSR
        embeddings_gen, labels_gen = LSR_gen(stats_for_LSR)
        embeddings_vis = concatenate(
            embeddings_vis,
            embeddings_gen)
        labels = concatenate(one_hot(labels), labels_gen)

    prototypes_text = compute_text_prototypes()
    logits = forward_with_bimodal_calibration(
        embeddings_vis,
        prototypes_text)
    return logits, labels

def forward_vision(images):
    embed = clip.visual.forward(images)
    shift_cur_domain = shifts[domain_id]
    embed = embed + c1 * shift_cur_domain
    if domain_id > 0:
        shift_base = shifts[0]
        embed = embed + c2 * shift_base.detach()
    embed = power_norm(embed, c3)

def compute_text_prototypes():
    prot_real_names = compute_prototypes_from_templates(
        class_names_list)
    prot_syn = compute_prototypes_from_templates(syn_list)
    prot_syn_weighted = Tensor()
    for name in class_names:
        sim = similarity(
            prot_real_name,
            prot_syn[class_name])
        weights = Softmax(sim)
        prot_weighted = multiply(weights, prot_syn)
        prot_syn_weighted = concatenate(
            prot_syn_weighted,
            prot_weighted)
    return interpolate(prot_syn_weighted, prot_real_names, c4)

def compute_prototypes_from_templates(names):
    texts = put_names_in_all_templates()
    embed_all_templates = forward_text(texts)
    embed_each_name = mean_for_templates(embed_all_templates)

def forward_text(texts):
    embed = clip.text_enc.forward(texts)
    shift_cur_domain = shifts[domain_id]
    embed = c5 + shift_text_vector + c6 * shift_cur_domain
    if domain_id > 0:
        shift_base = shifts[0]
        embed = embed + c7 * shift_base.detach()
    embed = power_norm(embed, c8)
  \end{lstlisting}
\end{algorithm}

\begin{algorithm}[htbp]
  \caption{PyTorch style pseudo-code for the LSR}
  \label{alg:LSR}
  \definecolor{codeblue}{rgb}{0.25,0.5,0.5}
  \lstset{
      basicstyle=\fontsize{7.2pt}{7.2pt}\ttfamily\bfseries,
      commentstyle=\fontsize{7.2pt}{7.2pt}\color{codeblue},
      keywordstyle=\fontsize{7.2pt}{7.2pt},
      language=python,
      breaklines=true,
      frame=single,
      aboveskip=0pt,
      belowskip=0pt,
    }
  \begin{lstlisting}[language=python]

def LSR_gen(stats):
    means, covariances, labels = mix_distributions(
    stats, num_candidates)

    means, covariances, labels = novel_novel_filter(
        means, covariances, labels,
        LSR_num_ways1)
    means, covariances, labels = novel_original_filter(
        means, covariances, labels,
        stats, LSR_num_ways2)

    embed_gen, labels_gen = Tensor(), Tensor()
    for i in range(LSR_num_ways2):
        pseudo_embed = sample_from_Gaussian(
        means[i],
        covariances[i],
        num_shots)
        embed_gen = concat(embed_gen, pseudo_embed)
        labels_gen = concat(
        labels_gen, labels[i].repeat(num_shots))
    
    return embed_gen, labels_gen
    
def mix_distributions(stats, num_candidates):
    means_gen, cov_gen, labels_gen = Tensor(), Tensor(), Tensor()
    for _ in range(num_candidates):
        mean1, cov1, lbl1, mean2, cov2, lbl2 = \
            sample_random_pairs(stats)
        c = sample_beta_dist()
        means_mixed = c * mean1 + (1-c) * mean2
        cov_mixed = c * cov1 + (1-c) * cov2
        label_mixed = c * lb1 + (1-c) * lb1
        means_gen = concat(means_gen, means_mixed)
        cov_gen = concat(cov_gen, cov_mixed)
        labels_gen = concat(labels_gen, label_mixed)
    return means_gen, cov_gen, labels_gen

def novel_novel_filter(
    means, covariances, labels,
    LSR_num_ways1
):
    sim = similarity(means, means)
    sim = remove_self_similarity(sim)
    score = sum(sim)
    indices = k_smallest(LSR_num_ways1, LSR_num_ways1)
    return means[indices], covariances[indices], label[indices]

def novel_original_filter(
    means, covariances, labels,
    stats_base,
    LSR_num_ways2
):
    means_base, cov_base, labels_base = stats_base.get_data()
    diversities = KL_div(
    means_base, cov_base,
    means, covariances
    )
    indices = k_largest(diversities, LSR_num_ways2)
    return means[indices], covariances[indices], label[indices]
  
  \end{lstlisting}
\end{algorithm}

\section{Complexity Analysis}
CLIP~\cite{Radford2021LearningTV} has two transformer branches: a vision encoder and a text encoder. The ViT branch has the most layers. Therefore, the time complexity is nearly twice that of the largest branch, and we only need to calculate the complexity of a Transformer in Big-O notation. \par

The most complex operations in a Transformer happen in the Multi-Head Self-Attention (MHSA) and Multi-Layer Perceptrons (MLPs). The patch embedding layer extracts $n=\frac{HW}{P^2} + 1$ from an image with height $H$, width $W$, and patch size of $P$. The same number of tokens is being processed in all $L$ layers of a transformer. In what follows, we first calculate the time complexity of a standard transformer. Next, we calculate the time complexity by considering the DCP. \par

\begin{figure*}[htbp]
   \centering
   \includegraphics[width=\textwidth]{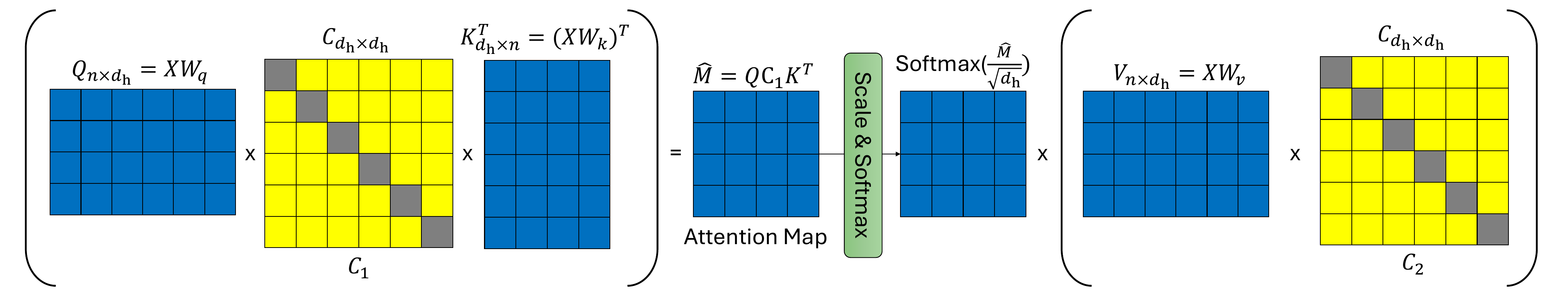}
   \caption{The calculations in the self-attention using the DCP.}
   \label{fig:DCP}
\end{figure*}

The DCP calculations are shown in \cref{fig:DCP}. For $d_h = \frac{d}{n_h}$, $d$ is the dimension of embeddings (width of the network), $n_h$ is the number of attention heads, and $d_h$ is the dimension of each head, the three projections of $Q$, $K$, and $V$ need $O(nd^2)$ operations. The $QK^T$ in a standard self-attention, costs $O(n_h n^2 d_h)=O(n^2d)$.
Additional multiplication to $V$ needs $O(n_h n^2 d_h)=O(n^2d)$. At last, the MLP for projecting the output needs $O(L_{M} n d d_{\text{M}})$, where $L_{M}$ is the number of MLP layers and $d_{M}$ is the dimension of hidden layers. As MLP is usually shallow (1 or 2 layers), the complexity becomes $O(n d d_{M})$. Overall, a block has the complexity of $O(nd^2 + n^2d + nd d_{\text{MLP}})$ \cite{paeedeh2025cross}. \par

In DCP, there are two CPs. For $C_1$ in the $QC_1K^T$ in Eq. 5 of the main paper, the $QC_1$ is $O(n_h n d_h^2) = O (\frac{nd^2}{n_h})$. The multiplication to the $K$ requires $O(n_h n^2 d_h)=O(n^2d)$. Therefore, the input of the Softmax needs $O(\frac{nd^2}{n_h} + n^2d)$. The $V C_2$ needs $O(n_h n d_h^2)$.
Therefore, the MHSA block with DCP requires the same $O(nd^2 + n^2d + nd d_{M})$ like the vanilla MHSA block, as the additional terms are asymptotically smaller than or equal to the other calculations, and absorbed in the Big-O notation. \par

In total, the time complexity of a Transformer with $L$ blocks stays $O(L (dn^2 + nd^2 + ndd_M))$.

\section{Evaluation metrics}

To evaluate our method, we followed the PGO-BEn~\cite{mukherjee2026textscpgoben} and used Overall Average Accuracy $AA^*$ and Overall Forgetting Alleviation $FA^*$. If we have trained the network from the domain $\mathcal{D}_1$ domain $\mathcal{D}_t$, we define Average Accuracy ($AA_t$) at domain $\mathcal{D}_t$ as follows:

\begin{equation}
    AA_t = \frac{1}{t} \sum_{i=1}^{t} A_{t,i},
\end{equation}
where $A_{t,i}$ indicates the accuracy of the model on $i$-th domain. $\text{AA}^*$ is defined as:

\begin{equation}
    AA^* = \frac{1}{T} \sum_{t=1}^{T}  AA_t.
\end{equation}

Moreover, if we have trained the network from the domain $\mathcal{D}_1$ to domain $\mathcal{D}_t$, where $t > 1$, for domain $\mathcal{D}_t$, Forgetting Alleviation for domain $\mathcal{D}_j$ is defined as:

\begin{equation}
    FA_j = \frac{1}{t - j} \sum_{i=j+1}^{t} A_{i,j}.
\end{equation}

The Overall Forgetting Alleviation $FA^*$ is defined by averaging the forgetting alleviations over all encountered domains ($T$) as follows:

\begin{equation}
    FA^* = \frac{1}{T-1} \sum_{t=1}^{T-1} FA_t
\end{equation}

\section{Experiments}

\subsection{More implementation details}

All experiments are performed with the ViT-B/16 variant of the CLIP for fair comparisons on an NVIDIA RTX 4090 GPU with 24 GB VRAM. In our experiments, we use 10 synonyms for each class name. AdamW~\cite{loshchilov2017decoupled} is used as the optimizer. We have utilized GPT-5.4~\cite{openai2026gpt54} to obtain 10 synonyms for each class name. For the CORe50 datasets, since there are 50 classes but 5 variants of the 10 classes (e.g., "Cup 1", "Cup 2", $\dots$, "Cup 5"), and all samples are just different frames from a short video clip of the same object, we added short descriptions of the objects as class names with Gemini 3.1 Pro and Flash \cite{google2026gemini31} and paraphrased them with GPT-5.4 as equivalents to the synonyms. \par

Following the PGO-BEn~\cite{mukherjee2026textscpgoben} experiments, the domains are encountered in each dataset as follows:
\begin{itemize}
    \item CDDB-Hard: 'GauGAN', 'BigGAN', 'Wild Deepfake', 'Which Face is Real', 'SAN'\\
    \item CORe50: 's1', 's2', 's3', 's4', 's5', 's6', 's7', 's8'\\
    \item DomainNet: 'Real', 'Painting', 'Clipart', 'Sketch', 'Quickdraw', 'Infograph'\\
\end{itemize}

\begin{table*}[htbp]
    \centering
    \resizebox{\linewidth}{!}{
    \begin{tabular}{c|cc|cc|cc}
    & \multicolumn{2}{c}{CDDB} & \multicolumn{2}{c}{CORe50} & \multicolumn{2}{c}{DomainNet} \\
        \toprule
    Name & Base Task & Inc. Tasks & Base Task & Inc. Tasks & Base Task & Inc. Tasks \\
        \toprule
        Num epochs                           & 5    & 8    & 4    & 8    & 2    & 5 \\
        Default lr.                          & 1e-3 & 1e-3 & 1e-3 & 1e-3 & 1e-3 & 1e-3 \\
        Lr. for PowerNorm                    & 2e-4 & 2e-4 & 2e-4 & 2e-4 & 2e-4 & 2e-4 \\
        Weight decay                         & 2e-5 & 2e-5 & 2e-5 & 2e-5 & 2e-5 & 2e-5 \\
        STD for CPs                          & 0.02 & 0.02 & 0.02 & 0.02 & 0.02 & 0.02 \\
        Num. candidates for LSR              & 100  & --   & 100  & -- & 300    & --   \\
        Num. generated pseudo-emb. per class & 10   & 10   & 10   & 10   & 10   & 10   \\
        Num. classes after novel-novel       & 8    & --   & 8    & --   & 24   & -- \\
        Num. classes after novel-original    & 4    & --   & 4    & --   & 12   & -- \\
        Num. synonyms                        & 10   & 10   & 10   & 10   & --   & -- \\
        Beta for LSR                         & 1    & --   & 1    & --   & 1    & -- \\
    \end{tabular}
    }
    \caption{The hyperparameters used in all experiments}
    \label{tab:hyper-parameters}
\end{table*}

\noindent We initialize each CP matrix in the DCP as follows:


\begin{equation}
    W_{ij} = \delta_{ij} + (1 - \delta_{ij})x_{ij}, \quad x_{ij} \sim \mathcal{N}(0, \sigma),
\end{equation}
Where $\delta_{ij}$ is the Kronecker delta function. The $\sigma=0.02$ in all experiments. Finally, the first 8 layers in both the vision and text encoders are domain-shared, and the last 4 layers are domain-specific. \cref{tab:hyper-parameters} shows the hyper-parameters used in our experiments.

\subsubsection{Analysis of the 5-shot experiments}

In addition to the 1-, 2-, 4-, and 8-shot settings, we compared our method with FSHIL~\cite{Li_2026_CVPR} in the 5-shot setting, using average and last accuracies. The comparison to the FSHIL, ASP~\cite{liu2024few}, App~\cite{wang2024approximation}, SEC~\cite{liu2025sec}, and ICON~\cite{park2024versatile} are provided in \cref{tab:FSHIL_5-shot}. Moreover, we tested our method with 3 random seeds. The results show that CVLC has more than a 10\% advantage in average accuracy on CORe50 and 22\% on DomainNet. The 9\% accuracy advantage on CORe50 and 22\% advantage on DomainNet after the last domains show that CVLC maintains the advantage in continual learning. Since CORe50's accuracy is comparable to that of PGO-BEn~\cite{mukherjee2026textscpgoben}, these results indicate that the advantages are consistent. \par

\begin{table}[htbp]
    \centering
    \resizebox{\linewidth}{!}{
    \begin{tabular}{cccccc}
        \toprule
            Method & Backbone & \multicolumn{2}{c}{CORe50} & \multicolumn{2}{c}{DomainNet} \\
            \cmidrule(lr){3-4} \cmidrule(lr){5-6}
            & & Avg.(↑) & Last(↑) & Avg.(↑) & Last(↑) \\
        \midrule
             ASP  \cite{liu2024few}            & ViT   & 85.94 & 84.33 & 45.43 & 16.47 \\
             App  \cite{wang2024approximation} & ViT   & 84.13 & 84.64 & 49.13 & 21.37 \\
             SEC  \cite{liu2025sec}            & ViT   & 84.72 & 81.82 & 47.53 & 18.19 \\
             ICON \cite{park2024versatile}     & ViT   & 65.82 & 75.59 & 32.56 & 32.27 \\
             FSHIL \cite{Li_2026_CVPR}         & ViT   & 86.61 & 87.01 & 52.92 & 44.05 \\
        \midrule
             CVLC                             & CLIP  & \valb{96.87}{0.09} & \valb{96.04}{0.23} & \valb{75.03}{0.16} & \valb{66.13}{0.35} \\
                                         & $\Delta$ & +10.20 & +9.03  & 22.11 & +22.08 \\
        \bottomrule
    \end{tabular}
    }
    \caption{Comparing the CVLC with other methods with 5-shot. The mean and standard deviations of the CVLC accuracies are reported across 3 random seeds.}
    \label{tab:FSHIL_5-shot}
\end{table}

In the 5-shot setting in \cref{tab:FSHIL_5-shot}, the average accuracy across all tasks is reported as follows:

\begin{equation}
    \text{Acc}_{\text{Avg.}} = \frac{1}{T} \sum_{t=1}^{T} \text{Acc}_{t},
\end{equation}
where $\text{Acc}_{i}$ is the accuracy on the test set after domain $\mathcal{D}_t$ is encountered, which is the proportion of correctly predicted labels to the total test samples. Finally, the last accuracy is $\text{Acc}_{T}$, which shows the final performance after the network is trained incrementally across all domains.

\subsubsection{Classes with insufficient samples}

In the cleaned version of the DomainNet, the following set of 39/345 classes have less than 8 samples at least in on of the domain during the training: \{ "angel", "shovel", "sleeping bag", "snail", "belt", "stereo", "stitches", "blackberry", "stove", "leaf", "light bulb", "syringe", "line", "bush", "calculator", "toe", "microwave", "cannon", "toothbrush", "t-shirt", "underwear", "van", "oven", "compass", "computer", "washing machine", "cooler", "waterslide", "pants", "cow", "wristwatch", "crown", "diamond", "diving board", "dresser", "eraser", "fan", "fireplace", "floor lamp" \}. Therefore, we exclude them in DomainNet experiments to be able to compare the experiments with different numbers of shots from the same classes. From this set, 9/39 classes have no training samples in at least one domain: \{ "toe", "t-shirt", "underwear", "syringe", "waterslide" , "floor lamp", "microwave", "toothbrush", "stereo" \}.

\subsection{UMAP Analysis}

\begin{figure}[htbp]
    \centering
    \captionsetup[subfigure]{justification=centering}

    \begin{subfigure}[b]{0.45\linewidth}
        \centering
        \includegraphics[width=\linewidth]{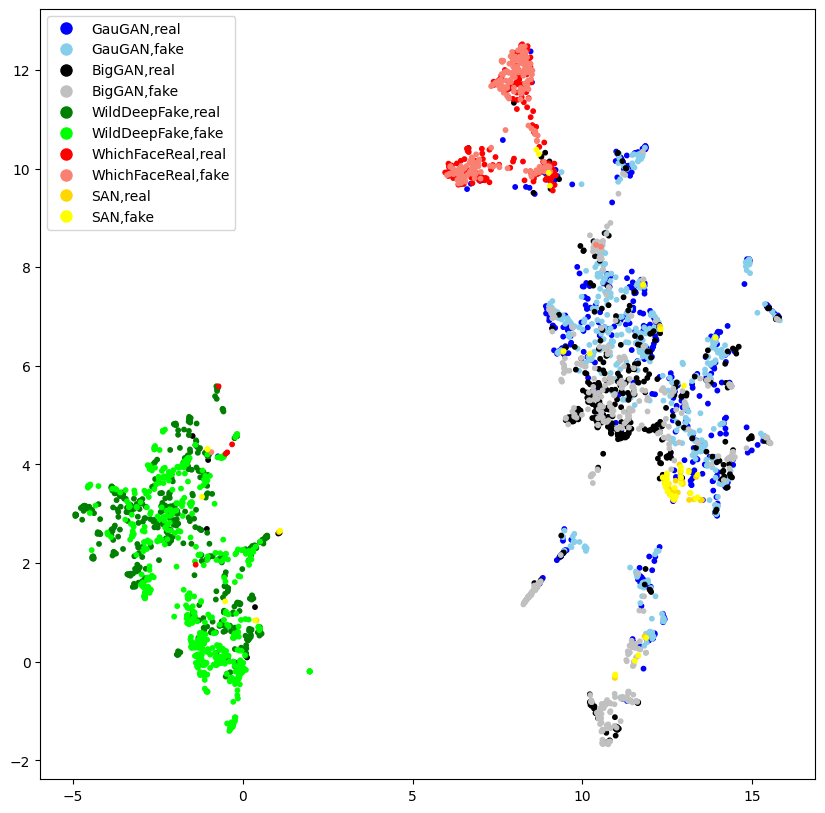} 
    \end{subfigure}
    \hfill
    \begin{subfigure}[b]{0.45\linewidth}
        \centering
        \includegraphics[width=\linewidth]{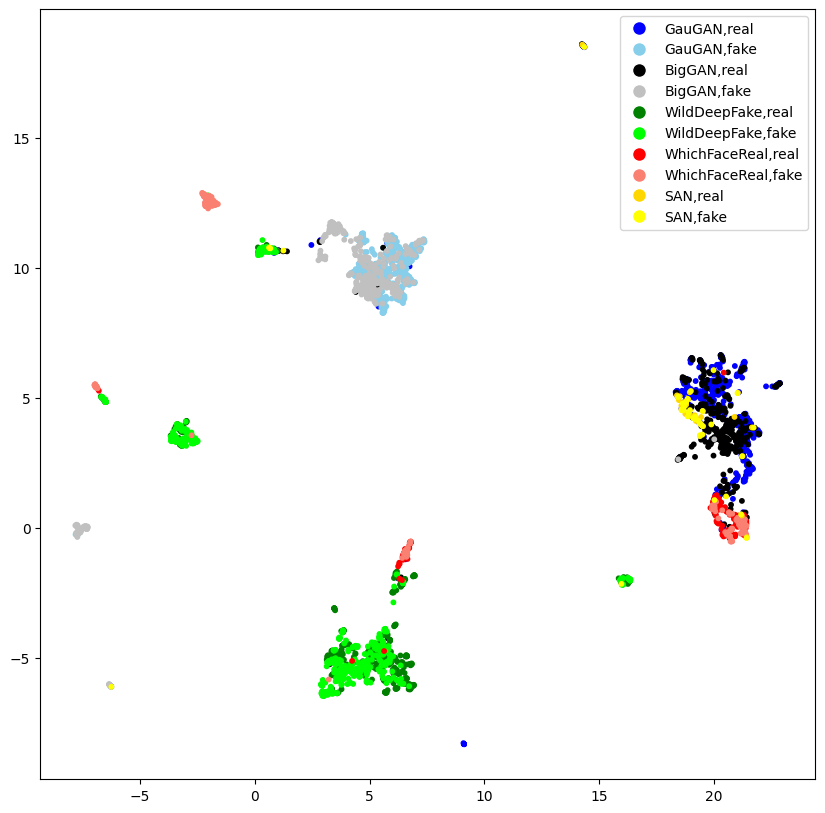}
    \end{subfigure}

    \caption{The UMAP plot of the embeddings, before training on the left and after training on the right, for all classes of the CDDB-Hard dataset.}
    \label{fig:UMAP}
\end{figure}

\cref{fig:UMAP} shows the UMAP plot of the embeddings for the 4-shot setting on the CDDB-Hard dataset, before and after training with CVLC. As the graph shows, the embeddings are more scattered before training, and the network is not expected to reliably distinguish the classes, since the classes in each domain are less separable. After training, the network's discriminative power improves, the embeddings become more organized, and the classes are clustered better. Moreover, the GauGAN and BigGAN classes remained well-separated despite being encountered first, indicating that the network is also robust to catastrophic forgetting.
\begin{figure}[htbp]
   \centering
   \includegraphics[width=\linewidth]{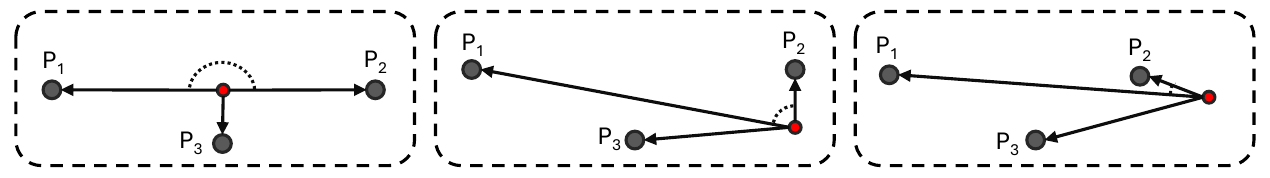}
   \caption{The effect of the shift in the location of the embeddings in a Cosine classifier. The center is shown with a red circle. The angle between $P_1$ and $P_2$ is highlighted with an arc for comparison. While the Cosine classifier is invariant to the magnitude of the vectors, the angles between the vectors can be dramatically changed.}
   \label{fig:shift}
\end{figure}

\section{Displacement of the embeddings}

One factor that is often neglected and affects the classifier's decisions is that shifts in the embeddings and prototypes can alter the magnitude and all angles between the embedding vectors or between the embeddings and prototypes, as shown in \cref{fig:shift}. While a Cosine classifier is invariant to the magnitudes of the vectors, the angles, hence Cosine similarity or Cosine distances are affected. Therefore, we consider learnable shifting vectors for each domain for the vision modality and a single vector for the text modality to find the optimum displacement for each modality. We assume that a shift vector learned on the base domain, along with a separate shift vector for each domain, can compensate for shifts in the embeddings, without requiring unnecessary backpropagation through the network's block. For the text encoder, we use an additional single vector to capture the shift in the text embeddings while also accounting for its partial influence from the calculated shifts in the base and current domains. More details about how these shift vectors are utilized are provided in the \cref{alg:pseudo-code,alg:LSR}.

\end{document}